\documentclass{article}
\usepackage{arxiv}
\usepackage{diagbox}
\usepackage[utf8]{inputenc} 
\usepackage[T1]{fontenc}    
\usepackage{hyperref}       
\usepackage{url}            
\usepackage{booktabs}       
\usepackage{amsfonts}       
\usepackage{nicefrac}       
\usepackage{microtype}      
\usepackage{lipsum}
\usepackage{graphicx}
\usepackage{xspace}
\usepackage{subcaption}
\usepackage{arydshln}

\usepackage[utf8]{inputenc} 
\usepackage[T1]{fontenc}    
\usepackage{hyperref}       
\usepackage{url}            
\usepackage{booktabs}       
\usepackage{amsfonts}       
\usepackage{nicefrac}       
\usepackage{microtype}      
\usepackage{xcolor}         
\usepackage{graphicx}
\usepackage{subcaption}
\usepackage{amsmath}
\usepackage{amsthm}
\usepackage{amssymb}
\usepackage{adjustbox}
\usepackage{wrapfig}
\usepackage[normalem]{ulem}
\usepackage{mathrsfs}
\usepackage{multirow}
\usepackage{svg}

\usepackage[linesnumbered,ruled,vlined]{algorithm2e}
\usepackage{soul}
\usepackage{enumitem}
\usepackage{pifont}
\usepackage{multirow}

\graphicspath{ {./figs-pdf/} }
\newcommand{\Lens}{\texttt{Lens}}

\title{\Lens: A Knowledge-Guided Foundation Model for Network Traffic}

\author{
\begin{tabular}{c}
Xiaochang Li\textsuperscript{1}\quad
Chen Qian\textsuperscript{1}\quad
Qineng Wang\textsuperscript{2}\quad
Jiangtao Kong\textsuperscript{1}\quad
Yuchen Wang\textsuperscript{1} \\
Ziyu Yao\textsuperscript{3}\quad
Bo Ji\textsuperscript{4}\quad
Long Cheng\textsuperscript{5}\quad
Gang Zhou\textsuperscript{1}\quad
Huajie Shao\textsuperscript{1*}
\end{tabular}
}

\begin{document}
\maketitle
\begingroup
\renewcommand\thefootnote{}\footnotetext{
\textsuperscript{*}Corresponding author.
\textsuperscript{1}William \& Mary;
\textsuperscript{2}Northwestern University;
\textsuperscript{3}George Mason University;
\textsuperscript{4}Virginia Tech;
\textsuperscript{5}Clemson University.
Correspondence to: Xiaochang Li<xli59@wm.edu>, Huajie Shao <hshao@wm.edu>.
}
\endgroup
\setcounter{footnote}{0}

\begin{abstract}
Network traffic refers to the amount of data being sent and received over the Internet or any system that connects computers. Analyzing network traffic is vital for security and management, yet remains challenging due to the heterogeneity of plain-text packet headers and encrypted payloads. To capture the latent semantics of traffic, recent studies have adopted Transformer-based pretraining techniques to learn network representations from massive traffic data. However, these methods pre-train on data-driven tasks but overlook network knowledge, such as masking partial digits of the indivisible network port numbers for prediction, thereby limiting semantic understanding. In addition, they struggle to extend classification to new classes during fine-tuning due to the distribution shift. Motivated by these limitations, we propose \Lens, a unified knowledge-guided foundation model for both network traffic classification and generation. In pretraining, we propose a Knowledge-Guided Mask Span Prediction method with textual context for learning knowledge-enriched representations. For extending to new classes in finetuning, we reframe the traffic classification as a closed-ended generation task and introduce context-aware finetuning to adapt the distribution shift. Evaluation results across various benchmark datasets demonstrate that the proposed \Lens~achieves superior performance on both classification and generation tasks. For traffic classification, \Lens~outperforms competitive baselines substantially on 8 out of 12 tasks with an average accuracy of \textbf{96.33\%} and extends to novel classes with significantly better performance. For traffic generation, \Lens~generates better high-fidelity network traffic for network simulation, gaining up to \textbf{30.46\%} and \textbf{33.3\%} better accuracy and F1 in fuzzing tests. We will open-source the code upon publication.
\end{abstract}


{
\section{Introduction}\label{sec:intro}
In computer networking, network traffic~\cite{oliveira2016computer} is defined as the flow of data, transmitted in the form of packets between interconnected computers or systems. The network packet consists of plain-text headers with metadata and an encrypted payload holding actual content. Given the critical role of networking, analyzing network traffic is crucial to ensure high network security, offer high-quality network services, and facilitate effective network management. 

Over the past decades, many approaches have been developed for network traffic analysis. Early works \cite{appscanner, bind, cumul, knn} mainly utilize statistical methods, heavily dependent on manually crafted features. To address this issue, some studies have employed deep learning methods~\cite{df, fs-net, deeppacket, datanet} to extract complicated features from raw data. While these approaches have shown impressive results in specific tasks, they often require extensive labeled data and struggle with generalization to new tasks. Consequently, recent works~\cite{etbert, yatc, netgpt, cui2025trafficllm} have developed foundation models employing various pretraining techniques to learn network representations from large-scale, raw traffic data. 


While they show promising results on various tasks, they still face two major \textit{limitations}. (i) Current data-driven pretraining methods~\cite{etbert, yatc} overlook vital network knowledge, resulting in incomplete semantic understanding. e.g., masking partial digits ``400'' of an indivisible port number ``40023'' for prediction. (ii) In finetuning for classification, most encoder-based models~\cite{pert, guthula2023netfound} use an additional multi-layer perceptron (MLP) as the classifier, which performs poorly when extending to new classes as they need to be re-trained for accommodation~\cite{li2017learning}. These limitations motivate us to address two fundamental research questions: \textbf{RQ1:} How can we integrate network-specific knowledge into pretraining to learn better network representations? \textbf{RQ2:} How can we seamlessly extend traffic classification to novel classes while maintaining high performance?

To answer these questions, we propose \Lens, a unified knowledge-guided foundation model based on an encoder-decoder T5 architecture~\cite{t5, codet5} to learn network representations from massive raw traffic data. The encoder-decoder architecture is chosen because it better captures the global information of the input data, making it well-suited for the header generation task based on its successor payload. In contrast, decoder-only models pretrained on next token prediction struggle with this due to their auto-regressive nature.
More specifically, to address \textbf{RQ1}, we propose the Knowledge-Guided Mask Span Prediction (KG-MSP) to intentionally mask network metadata and payload-related information as a whole based on their importance in networking. Plus, we incorporate context information as auxiliary knowledge into model pretraining.
For \textbf{RQ2}, we reframe the network traffic classification as a closed-ended generation task for adapting distribution shifts. Then, we leverage context-aware finetuning to smoothly extend classification from known to new classes by training only on new-class data with the updated context.

Finally, we assess the performance of \Lens~on 12 network traffic classification tasks and 5 network generation tasks across 6 datasets~\cite{vpn, tor, ustc, cp, doh, iot}. For the traffic classification, evaluation results show that \Lens~outperforms competitive baselines across 8 out of 12 tasks with an average accuracy of 96.31\%. Besides, \Lens~extends to classify novel classes well and gains accuracy and F1 advantage up to 31.31\% and 42.59\%, respectively. Regarding the traffic generation, \Lens~generates network traffic that more closely aligns with real-world data, achieving superior performance in network fuzzing tests with up to 30.46\% higher accuracy and a 33.3\% higher F1 score.

\noindent Our contributions are summarized as follows: 1) We propose \Lens, a novel knowledge-guided foundation model for network traffic. The proposed model integrates Knowledge-guided Mask Span Prediction with context for learning generalizable network representations in pretraining. 2) We reframe the traffic classification as a closed-ended generation task to adapt the distribution shift, which enables excellent extensibility by simply updating the textual context and lightly finetuning only on new classes. 3) We evaluate \Lens~on both network traffic classification and generation. For classification, \Lens~outperforms competitive baselines on most tasks and extends to novel classes significantly better. For generation, \Lens~generates high-fidelity network traffic that closely mirrors real-world distributions, enhancing the efficacy of subsequent fuzzing tests.

\section{Related Work} 
\subsection{Network Traffic Classification} 
\noindent\textbf{Classical Machine Learning Methods.} Earlier works have employed classical machine learning methods for network traffic analysis. For example, Wang et al.~\cite{knn} used the K-Nearest Neighbors (KNN) to identify attacks. CUMUL~\cite{cumul} adopted SVM for network traffic identification, while APPScanner and BIND~\cite{appscanner, bind} used statistical features like temporality and packet size to train Random Forests classifiers for identification tasks. Besides, IsAnon~\cite{isanon} fused Modified Mutual Information and Random Forest (MMIRF) to filter out redundant features. However, these methods require expert knowledge for feature extraction and lack generalization capability.

\noindent\textbf{Deep Learning Techniques.} Deep learning techniques have been introduced to provide a more automated approach to comprehend network traffic without human-designed features. DF~\cite{df}, for example, devised Convolutional Neural Networks (CNN) for identifying a novel website fingerprinting attack. In addition, FS-Net~\cite{fs-net, tscrnn} employed recurrent neural networks (RNN) and its variant LSTM~\cite{bilstm,datanet} to classify network traffic. Recently, a method called DeepPacket~\cite{deeppacket} has been introduced. It combined the stacked autoencoder (SAE) with CNN to identify and extract the important features for traffic classification tasks. However, these approaches rely heavily on large amounts of labeled data and have limited generalization ability.


\noindent\textbf{Pre-training Approaches.} To improve model generalization ability, recent studies have adopted pretraining techniques to learn representations from large-scale traffic data in an unsupervised manner. For instance, PERT~\cite{pert} and ET-BERT~\cite{etbert} leveraged the ALBERT~\cite{albert} and BERT~\cite{bert} to learn the latent network representations, respectively. Lately, netFound~\cite{guthula2023netfound} and NetMamba~\cite{wang2024netmamba} pre-trained hierarchical Transformers and state space models for learning network representations separately. However, these models pre-train on random masking and are only applicable for classification due to their encoder-only structure. At the same time, Decoder-based models like NetGPT~\cite{netgpt}, GBC~\cite{zhao2025language}, and TrafficGPT~\cite{qu2024trafficgpt}~\footnote{NetGPT, GBC and TrafficGPT are currently preprinted and not open-sourced, and we thus do not compare our method with them.} are pre-trained on the next token prediction for both traffic classification and generation. Nevertheless, they predict tokens based on causal probabilities, inferring the value of network fields based on partial context. More recently, works like TrafficLLM~\cite{cui2025trafficllm} fine-tuned LLM pretrained on natural language for traffic classification and generation. However, it suffers from hallucination and requires more labeled data to mitigate the domain gap. Different from prior works, we pre-train a network foundation model with a knowledge-guided task combined with auxiliary context to learn better network representations. Table~\ref{tab:diff} summarizes the comparison of the proposed \Lens~and other existing methods.

\begin{table}[!htb]
    \small
    \centering
    \caption{Comparison of the proposed \Lens ~and existing pretraining methods. ``KG Pretrain'' denotes knowledge-guided pretraining. ``IP Masking'' means all IPs are anonymized or removed in both pretraining and finetuning for privacy. ``Generation'' refers to the support of generation tasks.}
    \label{tab:diff}
    \begin{adjustbox}{width=0.8\textwidth}
    \begin{tabular}{cccccc}
        \toprule
        \textbf{Method} & \textbf{Encoder} & \textbf{Decoder} & \textbf{KG Pretrain} & \textbf{IP Mask} & \textbf{Generation}\\
        \midrule
        PERT \cite{pert} & \ding{52} & \ding{56} & \ding{56} & \ding{56} & \ding{56}\\
         ET-BERT \cite{etbert} & \ding{52} & \ding{56} & \ding{56} & \ding{52}& \ding{56}\\
         NetGPT \cite{netgpt} & \ding{56} & \ding{52} & \ding{56} & \ding{56} & \ding{52}\\
         YaTC \cite{yatc} & \ding{52} & \ding{56} & \ding{56} & \ding{52} & \ding{56}\\
         TrafficLLM \cite{cui2025trafficllm}  & \ding{56} & \ding{52} & \ding{56} & \ding{52} & \ding{52}\\
        \midrule
         \textbf{\Lens (Ours)} & \ding{52} & \ding{52} & \ding{52} & \ding{52}& \ding{52}\\
         \bottomrule
    \end{tabular}
    \end{adjustbox}
\end{table}

\subsection{Network Traffic Generation}
\noindent\textbf{Tool-Based Traffic Generation.} Classical traffic generation methods mainly focus on simulation tools and structure-based solutions. Simulation tools, such as NS-3~\cite{ns-3}, yans~\cite{yans}, and DYNAMO~\cite{dynamo}, are based on varying network topology. Structure-based methods like Iperf~\cite{botta2gentool}, Harpoon~\cite{harpoon}, and Swing~\cite{swing} capture network patterns via heuristics. However, these methods require vast domain expertise and might lack versatility. Moreover, tool-based traffic generation methods often result in rigid traffic patterns that may not accurately reflect the dynamic and stochastic nature of real-world network conditions. Thus, they cannot effectively adapt to the evolving behaviors of cyber threats or user demands.

\noindent\textbf{GAN-Based Traffic Generation.} In addition, some studies have employed Generative Adversarial Networks (GAN) to generate network traffic. Ring et al.~\cite{first-gan} first suggested the use of GAN~\cite{gan} for the simulation of flow-level traffic. The following works include NetShare~\cite{netshare} and DoppelGANger~\cite{DoppelGANger}, and others~\cite{gan2021, gan2022}. For instance, NetShare generated packet and flow header traces for networking tasks, such as telemetry, anomaly detection, and provisioning. Though GAN-based methods are adaptive, their generated results may be inconsistent with target protocols~\cite{netshare}.

\noindent\textbf{Pretraining-based Generation.} Recently, researchers employed a pretraining technique for traffic generation. The Decoder-based NetGPT~\cite{netgpt} and TrafficGPT~\cite{qu2024trafficgpt} have been developed to generate key network header fields. However, it is hard to assess their performance as they did not compare results with the state-of-the-art models like Netshare. More recently, NetDiffusion~\cite{jiang2024netdiffusion} applied diffusion models to generate network traffic, while Chu et al.~\cite{chu2024feasibility} investigated the feasibility of state space models in network traffic generation. Nevertheless, our models can adapt to both classification and generation. Most recently, TrafficLLM~\cite{cui2025trafficllm} fine-tuned LLM to generate network packets, but it requires more data and computational resources to achieve good performance. Unlike previous works, our \Lens~leverages context-aware finetuning to generate high-fidelity network traffic available for downstream simulation.

\begin{figure*}[!thb]
    \centering
    \includegraphics[width=\textwidth]{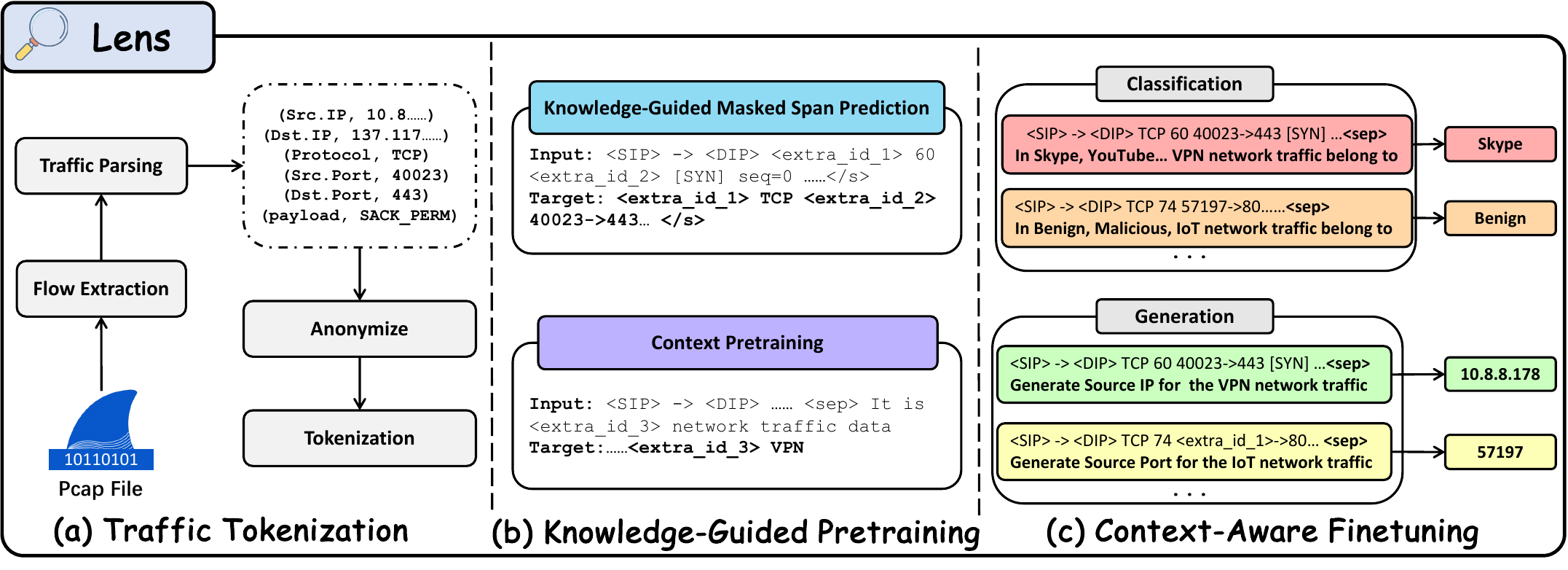}
    \caption{The overall framework of \Lens. (a) Network flows are extracted, parsed with Tshark, anonymized, and tokenized using our network-specific tokenizer. (b) \Lens~is pretrained with Knowledge-Guided Masked Span Prediction (KG-MSP) and auxiliary natural-language context. (c) In finetuning, \Lens~performs downstream classification and generation tasks via context-aware finetuning.}
    \label{fig:overview}
\end{figure*}
\section{Overall Framework of LENS}\label{sec:model}
We develop \Lens, a unified knowledge-guided foundation model for network traffic classification and generation. The Figure~\ref{fig:overview} illustrates the overall framework, consisting of three main stages: (a) traffic tokenization, (b) knowledge-guided pretraining, and (c) context-aware finetuning. In Section~\ref{sec:Ldpp}, we detail how we pre-process and conduct network-specific tokenization to benefit representation learning. Then, Section~\ref{sec:Lpt} presents the knowledge-guided Masked Span Prediction (KG-MSP) with the context pretraining. Lastly, Section~\ref{sec:finetune} demonstrates the context-aware finetuning on downstream network traffic classification and generation tasks. 

\subsection{Traffic Tokenization}\label{sec:Ldpp}
To preprocess input data with network knowledge, we parse network flows through Tshark~\cite{wireshark} into text-like input as shown in Figure~\ref{fig:dataexample} of the Appendix~\ref{appendix:input_example}. Afterwards, we pretrain a specialized Byte-level Byte Pair Encoding (BBPE)~\cite{bbpe} tokenizer on the textual network input to tokenize network terms, like ``Seq'', ``TCP'', and natural language context properly. More details about the data preprocessing are elaborated in the Appendix~\ref{appendix:datapreprocess}.


\subsection{Knowledge-guided Pretraining}\label{sec:Lpt}
We pretrain the proposed \Lens~using Knowledge-Guided Masked Span Prediction (KG-MSP), an objective that intentionally masks significant network metadata and payload-related information entirely to learn generalizable representations for downstream tasks. By also incorporating textual context as auxiliary knowledge, \Lens~is able to learn more generalizable representations after pre-training on both network traffic and natural language.

\subsubsection{Knowledge-guided Mask Span Prediction}\label{sec:msp}
\begin{figure*}[!thb]
    \centering
    \includegraphics[width=\textwidth]{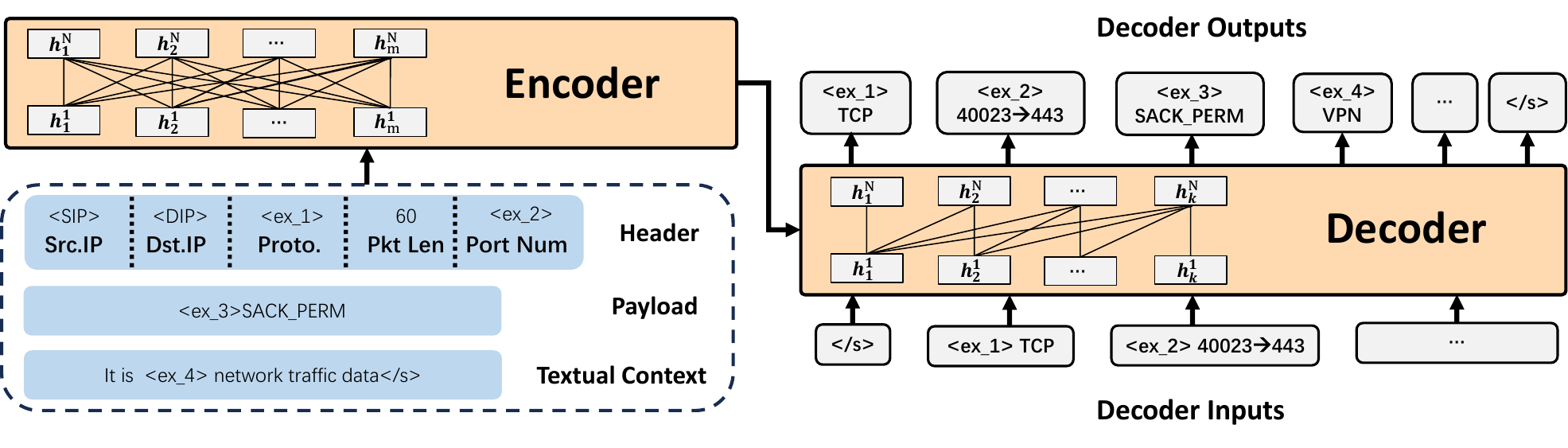}
    \caption{The core model architecture of \Lens~for pre-training with both the encoder and decoder. 1) The encoder takes in masked network traffic (header and payload) and textual template context. 2) The decoder uncovers the masked span tokens in both traffic and context based on Encoder representations in an auto-regressive way.}
    \label{fig:arch}
\end{figure*}

The data-driven random masking~\cite{bert, etbert} has been proved effective in pretraining across many NLP tasks. The basic idea is to first mask random tokens in the input and then recover them at the decoder side.  However, directly applying random masking may degrade network representation, since it may mask network metadata partially. For example, masking partial digits "400" of the port number "40023" for prediction harms the understanding of the port number as a complete semantic unit. Similarly, masking the "T" of "TCP" also hurts the representation of the network protocol. 

Motivated by this observation, we propose a new knowledge-guided masking method to intentionally mask vital network metadata such as protocol and port numbers entirely, thereby learning better network representations, as shown in Figure~\ref{fig:arch}. As mentioned before, the parsed network packet contains important network-specific knowledge in both the header and the payload. In both TCP and UDP network traffic, the header meta-information includes anonymized IP addresses, network protocol, packet length, port numbers, and payload length. To ensure network reliability, the TCP packet owns more header fields like sequence, acknowledgment numbers, maximum segment size, and so on. Besides, other text-like network messages like SACK\_PERM, Client Hello, indicate the purpose of the sent packet. To learn the generalizable representations of network metadata in pretraining, we set different probabilities to mask them based on their importance in networking. 

\textbf{For identifying application types, port numbers and protocols are critical.} For example, VPN applications typically use port 443 for secure TCP transmission to protect the content's safety. Besides, most applications use the DNS protocol for looking up the IP address of a network domain. Given its importance, we set a $\theta$ probability to mask them for learning insightful network representations empirically. Both the source and destination port numbers are masked for a better understanding of the network traffic direction. 

\textbf{For understanding packet intentions, protocol flags and textual payload messages are essential.}
For instance, the network flag ``[SYN]'' means synchronization of the TCP 3-way handshake, indicating the start of a TCP session. Similarly, the text-like TLS handshake message Server Key Exchange hides the process of exchanging encrypted keys but preserves the process of building connections. Since protocol flags and messages are essential for interpreting underlying interactions within network flows, we also set the same probability of $\theta$ to mask them as a whole. To find the $\theta$, we conducted a sensitivity analysis as Table~\ref{tab:sensitive_analysis} in the Appendix~\ref{appendix:sensitive}.

\textbf{For analyzing protocol behavior, sequence and acknowledgment numbers are fundamental.}
They record how many bytes are exchanged between the server and client, ensuring reliable TCP transmission. But, their patterns are comparatively dynamic as the retransmission process or the loss of packets in the network traffic changes them from time to time. Likewise, the packet and payload length vary for more efficient transmission in different network conditions. Thus, we assign a probability of $k$ to mask sequence numbers, acknowledgment numbers, and the length of packets and payloads. We conducted sensitivity analysis on choosing $k$ in Table~\ref{tab:sense_seq} in the Appendix~\ref{appendix:sensitive}. 

To preserve the randomness in the masking policy, we further mask randomly on the rest of the input to satisfy the overall 15\% masking ratio as~\cite{t5}. To calculate the loss function of KG-MSP, we use the following negative log likelihood
\begin{equation}
    \mathcal{L}_\text{KG-MSP} = -\sum_{i=1}^k\log P(\text{KSPAN}_i=\text{SPAN}_i|\mathbf{x}_\text{in}, \text{KSPAN}_{<i};\theta), \label{formula:kgmsp}
\end{equation}
where $\theta$ represents the model (both encoder and decoder) parameters, $k$ is the total number of masked spans, $\text{KSPAN}_i$ is $i$-th span generated by the decoder part of \Lens, $\text{SPAN}_i$ denotes the original span tokens, $\mathbf{x}_\text{in}$ is the masked input sequence of the encoder, $\text{KSPAN}_{<i}$ indicates the generated spans from the decoder before $i$-th span. $\text{KSPAN}_i$ is generated given $\mathbf{x}_\text{in}$ and $\text{KSPAN}_{<i}$ when parameterized by $\theta$. Here, \Lens~ decodes the original span tokens after the special token used for masking in input as ~\cite{t5}.

To demonstrate KG-MSP's superiority, we compare KG-MSP with random masking used in ET-BERT~\cite{etbert}. In essence, it masks overall 15\% of the input randomly on single tokens for prediction. We pretrain the random masking with \Lens's base model and evaluate on the downstream classification task. The ablation results in Table~\ref {table:mspablation} in the Appendix~\ref{appendix:msp} demonstrate the superiority of KG-MSP. 

\textbf{Context Pretraining. }Aside from learning better network representations via KG-MSP, we combine the textual description of network traffic input as auxiliary knowledge. As shown in Figure~\ref {fig:arch}, the textual description like \textit{It is VPN network traffic data} is appended after the network traffic with the special "<sep>" token in the middle separating the two modalities. The textual description is derived solely from the dataset source and does not contain any downstream task labels, ensuring that no task-specific information is leaked. The benefits are two-sided: 1) The auxiliary description helps differentiate similar network traffic via its sources, like VPN, DoHBrw, IoT, and so on. 2) The textual description bridges the gap between network traffic and natural language, preparing basic natural language understanding for downstream context-aware finetuning. To pretrain the auxiliary knowledge, we randomly mask 15\% of the textual description as NLP pretraining.
\subsection{Context-aware Finetuning}\label{sec:finetune} 
After knowledge-guided pretraining, we fine-tune the proposed \Lens~on various downstream network traffic classification and generation tasks. For the traffic classification, \Lens~reframes it as a closed-ended generation task and then leverages context-aware finetuning to generate the label at the decoder side. This reformation enables \Lens~to adapt distribution shifts easily and extend to classify new classes, which benefits generalization in fast-changing network environments. In addition, \Lens~can also be fine-tuned to generate high-fidelity network traffic for better downstream network simulation, such as fuzzing tests.

\textbf{Network Traffic Classification. }As shown in Figure~\ref{fig:overview} (c), the input starts with the parsed network traffic and the task context with the format: \textit{network traffic} <sep> \textit{In Skype, Youtube, $\dots$, the VPN network traffic belongs to}. In the task context, we provide all application labels for \Lens~to select. This provides potential for extensibility to unseen classes through adding more label options in the context. After feeding the network traffic and the task context to the encoder, \Lens~generates labels on the decoder side based on input representations. To prevent generating labels out of choices, we guarantee the validity of the output by using a prefix trie as GENRE~\cite{decao2021autoregressive} to constrain the generation process, ensuring the output sequence always matches a known label. 

\textbf{Extending Network Classification to Novel Classes. }Reforming traffic classification into a closed-ended generation task, \Lens~alleviates the distribution shifts and seamlessly extends to new classes via context-aware finetuning. Specifically, after encountering unseen data from new classes, \Lens~reloads the saved checkpoint to resume the learned knowledge on known classes first. Then, it adds new label options in the task context and only fine-tunes on unseen data, correlating new labels to unseen data. Through this process, \Lens~reuses learned knowledge for classifying new classes seamlessly through discerning the differences in the task context. 

\textbf{Network Traffic Generation. }For the generation task, the context-aware finetuning also benefits \Lens~significantly through providing auxiliary knowledge. As shown in Figure~\ref{fig:overview} (c), the input format for the generation task is like \textit{network traffic} <sep> \textit{Generate Source IP for the VPN network traffic}. To complete different generation tasks, we substitute the task context with various generation targets, such as port numbers, and mask the corresponding fields in the network traffic input for generation.
After generating necessary header fields, we evaluate the fidelity of synthetic network traffic on fuzzing tests~\cite{munea2016network} to verify its usefulness. 
Following NetShare~\cite{netshare}'s setting, generating encrypted payloads is a lower priority, as most real-world payloads are encrypted and uninformative. Instead, replaying captured payloads or applying standard encryption algorithms provides a more realistic and practical reconstruction of network traffic payload.


\section{Expeiment}\label{sec:experiment}
Extensive experiments are carried out to evaluate the performance of \Lens~on both network traffic classification and generation tasks across 6 different datasets. First, we elaborate on the implementation details and experimental settings, including datasets, baselines, and evaluation metrics. Then, in Section~\ref{sec:classification} and ~\ref{sec:extensibility}, we compare our \Lens~with competitive baselines on 12 classification tasks and demonstrate its extensibility to new classes. Afterwards, in Section~\ref{sec:generation}, we showcase \Lens~'s traffic generation capabilities and simulation results in fuzzing tests. Finally, we conduct ablation studies to further investigate the impact of the main components on model performance in Section~\ref{sec:ablation}.

\textbf{Datasets.} In our experiments, we use six publicly available datasets in NetBench~\cite{qian2024netbench}, including ISCX-VPN~\cite{vpn}, ISCX-Tor~\cite{tor}, USTC-TFC-2016~\cite{ustc}, Cross Platform \cite{cp}, CIC-DoHBrw-2020~\cite{doh}, and the CIC-IoT-2023 \cite{iot}. Based on these 6 datasets, we conduct 12 downstream network traffic classification and 5 network generation tasks. In the following, we will detail how we pre-process and organize pretraining and finetuning datasets. The detailed dataset distribution between pretraining and finetuning datasets are listed in Table~\ref{tab:ptftdist} in Appendix~\ref{appendix:ptftpercent}.

\noindent\textbf{Pretraining Data.} For each dataset (except CrossPlatforms), we randomly sample 60\% of flows using a stratified sampling strategy, but only with respect to the coarsest dataset-level categories (e.g., VPN vs non-VPN, benign vs malicious), without using any fine-grained task labels such as application or service types. This ensures that pretraining does not access any downstream classification labels and completely avoids label leakage. The pretraining split is also used to pretrain our network-specific BBPE tokenizer, together with the mini C4 natural language corpus~\cite{miranda2021ultimateutils}.

\noindent\textbf{Finetuning Data:} For finetuning, we randomly select around 10k network flows for finetune training for each dataset except CrossPlatform. Since downstream labels like "YouTube" and "Skype" are not trained in the pretraining stage, fine-tuning requires sufficient data to learn these label strings. Moreover, we preserve the imbalanced nature in both training and testing, like real-world situation. 

We conduct flow-level classification and packet-level generation. Specifically, \Lens~classifies each flow using its first 34 packets and performs packet-level generation using one packet sampled from each flow for universality. In the classification task, the choice of 34 packets is derived from the dataset statistics in Table~\ref{flowlenstats} of Appendix~\ref{appendix:datapreprocess}. In the generation task, we mask the header field that the model is required to predict in both training and testing. Representative data examples are shown in Figure~\ref{fig:dataexample} in Appendix~\ref{appendix:input_example}.

\textbf{Implementation Details.} We adopt Google's T5-v1.1-base model \cite{t5}, which contains roughly 0.25B parameters, as the foundational architecture for the proposed \Lens. To process the input length up to 1,500, we develop \Lens~based on the TurboT5's implementation~\cite{Knowledgator_TurboT5_2023}. All experiments are conducted on a GPU server with 4 NVIDIA A6000 48G GPUs, Ubuntu (20.04.6). For pretraining, we set the batch size to $48$ and implement gradient accumulation over $6$ training steps. The learning rate is set to $5\times10^{-4}$ over a total of 130,000 steps with 10\% steps as a warm-up. For finetuning, we set the batch size to 32, dropout rate to $0.1$, and the learning rate to $5\times10^{-5}$, training each downstream task for 20 epochs using the AdamW optimizer~\cite{adamw}. More details are detailed in Table~\ref{tab:hyperparameters} of the Appendix~\ref{appendix:hyper}.

\textbf{Baselines. }To compare the performance of \Lens, we compare with two types of baselines: (i) \texttt{Traffic Classification Baselines:} We firstly compare against the deep learning baselines, including FS-Net~\cite{fs-net}, BiLSTM\_Att~\cite{bilstm}, Datanet~\cite{datanet}, DeepPacket~\cite{deeppacket}, and TSCRNN~\cite{tscrnn}. Then, we compare against competitive pretraining-based methods, including YaTC~\cite{yatc}, TrafficLLM~\cite{cui2025trafficllm}, and ET-BERT~\cite{etbert}. (ii) \texttt{Traffic Generation Baselines:} For network generation, we compare against the GAN-based Netshare~\cite{netshare} and the pretraining-based TrafficLLM~\cite{cui2025trafficllm}.

\textbf{Evaluation Metrics. }For the network traffic classification tasks, the Accuracy (AC) and Macro F1 Score (F1) are used to assess the performance. For the netowrk traffic generation, the Jensen-Shannon Divergence (JSD) and Total Variation Distance (TVD) are used, following NetGPT~\cite{netgpt}'s experimental setting. The JSD measures how similar two probability distributions are, while the TVD identifies the largest difference in probabilities between two distributions. Lower JSD and TVD indicate better performance.

\subsection{Network Traffic Classification Performance}\label{sec:classification}
We evaluate \Lens's traffic classification capabilities against two lines of state-of-the-art approaches: deep learning models and pretraining-based foundation models in Table~\ref{tab:compUnder1} and \ref{tab:compUnder2}. All deep learning and pretraining-based models are implemented on open-sourced code and follow the experimental setting as described in their paper. The 12 classification tasks are detailed as below: For ISCX-VPN, we have VPN detection (Task 1), VPN Service detection (Task 2), and VPN application classification (Task 3). For the ISCX-Tor and USTC-TFC-2016, our task involves Tor service detection (Task 4) and USTC-TFC-2016 Application Classification (Task 5). For the Cross Platform (Android), we have an application classification (Task 6) and a country detection task (Task 7). Similarly, for the Cross Platform (IOS), we have application classification (Task 8) and country detection (Task 9). For CIC-DoHBrw-2020(DoH), we include DoH query method classification (Task 10). In the CIC-IoT-2023, we have the IoT attack detection (Task 11) and IoT attack method detection (Task 12). More details about the task and class numbers are detailed in Table~\ref{tab:tasks} in Appendix~\ref{appendix:tasks}.

\begin{table*}[!thb]
    \centering
    \caption{Performance on traffic classification from Task 1 to Task 6. \Lens achieves superior results than baselines, especially on Task 2, 3, and 4. Bold numbers denote the best results, while underlined numbers are the second best. }
    \label{tab:compUnder1}
    \begin{adjustbox}{width=\textwidth}
    \begin{tabular}{l  c c  c c  c c  c c  c c  c c }
        \toprule
        \multirow{2}{*}{\centering Method} &
        \multicolumn{2}{c}{Task 1} &
        \multicolumn{2}{c}{Task 2} &
        \multicolumn{2}{c}{Task 3} &
        \multicolumn{2}{c}{Task 4} &
        \multicolumn{2}{c}{Task 5} &
        \multicolumn{2}{c}{Task 6} \\
        \cmidrule{2-13}
        & AC & F1 & AC & F1 & AC & F1 & AC & F1 & AC & F1 & AC & F1 \\
        \midrule
        FS-Net & 0.9785 & 0.9537 & 0.7360 & 0.6732 & 0.5681 & 0.5910 & 0.9271 & \underline{0.7606} & 0.8074 & 0.8817 & 0.2822 & 0.1219 \\
        BiLSTM\_Att & 0.9798 & 0.9551 & 0.8009 & 0.7719 & 0.6103 & 0.6635 & \underline{0.9421} & 0.6095 & 0.9463 & 0.9568 & 0.8091 & 0.6023\\
        Datanet & 0.9726 & 0.9405 & 0.7755 & 0.7209 & 0.5762 & 0.5717 & 0.9362 & 0.5483 & 0.9397 & 0.9540 & 0.7081 & 0.4566 \\
        DeepPacket & 0.9603 & 0.9178 & 0.7934 & 0.7585 & 0.6137 & 0.6834 & 0.9410 & 0.6405 & 0.9372 & 0.9456 & 0.7993 & 0.5622 \\
        TSCRNN & 0.9668 & 0.9233 & 0.7975 & 0.7568 & 0.6028 & 0.6483 & 0.9368 & 0.6049 & 0.9412 & 0.9513 & 0.9072 & 0.8319 \\
        \midrule
        YaTC & 0.9834 & 0.9627 & 0.8078 & 0.7805 & 0.6420 & \underline{0.6990} & 0.9362 & 0.6956 & \underline{0.9533} & \textbf{0.9707} & 0.9409 & 0.8173 \\
        TrafficLLM & 0.9083 & 0.8084 & 0.6757 & 0.4826 & 0.5092 & 0.4650 & 0.9415 & 0.7076 & 0.6598 & 0.6363 & NA & NA \\
        ET-Bert & \underline{0.9863} & \underline{0.9692} & \underline{0.8130} & \underline{0.8033} & \underline{0.6484} & 0.6662 & 0.9296 & 0.6336 & 0.9462 & 0.9604 & \textbf{0.9800} & \textbf{0.8925}\\
        \midrule
        \Lens~(Ours) & \textbf{0.9942} & \textbf{0.9870} & \textbf{0.8979} & \textbf{0.8893} & \textbf{0.8406} & \textbf{0.8137} & \textbf{0.9692} & \textbf{0.8120} & \textbf{0.9538} & \underline{0.9676} & \underline{0.9660} & \underline{0.8847} \\
        \bottomrule
    \end{tabular}
    \end{adjustbox}
\end{table*}

\textbf{In both Table~\ref{tab:compUnder1} and~\ref{tab:compUnder2}, we can see that \Lens~outperforms all baselines with superior accuracy and F1 in 8 out of 12 tasks while performing comparably in the remaining tasks.} The results indicate that pretraining-based models mostly outperform deep learning models, demonstrating the benefits of pretrained network representations. The TrafficLLM~\cite{cui2025trafficllm} does not perform well, as it suffers from the modality gap between network input and natural language, but also the imbalanced label distribution. This shows the necessity of pre-training network foundation models on network traffic data. On challenging Tasks 2, 3, 4, 10, and 12, \Lens~outperforms all other baselines significantly. In ISCX-VPN datasets (Task 2 and Task3), \Lens~achieves accuracies of 84.06\% and 89.79\% , representing a +8.49\% and +20.64\% accuracy improvement over YaTC and ET-BERT separately. The F1 improvement of +11.47\% on Task 2 and +8.49\% on Task 3 further demonstrate \Lens's robust performance on the imbalanced test set. For ISCX-Tor dataset(Task 4), DoH dataset(Task 10), and IoT (Task 12), \Lens~surpasses the second-best baselines significantly by +11.13\%, +5.26\%, +9.17\% on F1, while achieving higher accuracy than baselines on all these tasks. The main reason is that \Lens's pretraining on KG-MSP with the context has learned generalizable network representations, better capturing networking semantics. Besides, the context-aware finetuning provides auxiliary task information and label options, finetuning pretrained representations to output labels well. 

\begin{table*}[thb]
    \centering
    \caption{Performance on traffic classification from Tasks 7 to 12. \Lens~outperforms baselines significantly on Task 10 and 12, while achieves excellent generalization capabilities. ``Avg.'' is the average accuracy. }
    \label{tab:compUnder2}
    \begin{adjustbox}{width=\textwidth}
    \begin{tabular}{l  c c  c c  c c  c c  c c  c c  c }
        \toprule
        \multirow{2}{*}{\centering Method} &
        \multicolumn{2}{c}{Task 7} &
        \multicolumn{2}{c}{Task 8} &
        \multicolumn{2}{c}{Task 9} &
        \multicolumn{2}{c}{Task 10} &
        \multicolumn{2}{c}{Task 11} &
        \multicolumn{2}{c}{Task 12} &
        Avg.\\
        \cmidrule{2-14}
        & AC & F1 & AC & F1 & AC & F1 & AC & F1 & AC & F1 & AC & F1 & AC\\
        \midrule
        FS-Net & 0.8552 & 0.5157 & 0.2457 & 0.1052 & 0.3605 & 0.1767 & 0.5275 & 0.1381 & 0.3808 & 0.2758 & 0.9448 & 0.4270 & 0.6345\\
        BiLSTM\_Att & 0.9440 & 0.8519 & 0.9080 & 0.8414 & 0.9563 & 0.9567 & 0.9781 & 0.7024 & 0.9809 & 0.9798 & 0.9695 & 0.4283 & 0.9021\\
        Datanet & 0.9141 & 0.7855 & 0.6527 & 0.4508 & 0.9343 & 0.9347 & 0.9742 & 0.6564 & 0.9799 & 0.9788 & 0.9454 & 0.4518 & 0.8591\\
        DeepPacket & 0.9291 & 0.8210 & 0.6330 & 0.3435 & 0.9412 & 0.9414 & 0.9774 & 0.7321 & 0.9814 & 0.9804 & 0.9655 & 0.4714 & 0.8727\\
        TSCRNN & 0.9267 & 0.7953 & 0.9072 & 0.8319 & 0.9510 & 0.9513 & 0.9781 & 0.7024 & 0.9800 & 0.9790 & 0.9698 & 0.4604 & 0.9054\\
        \midrule
        YaTC & \underline{0.9960} & \underline{0.9896} & 0.9552 & 0.9160 & \underline{0.9963} & \underline{0.9962} & \underline{0.9905} & \underline{0.9083} & \underline{0.9864} & \underline{0.9857} &\underline{ 0.9586} & \underline{0.5885} & 0.9289\\
        TrafficLLM & 0.9473 & 0.8519 & NA & NA & 0.9865 & 0.9864 & 0.4680 & 0.2461 & 0.7927 & 0.7439 & 0.5472 & 0.1844 & 0.7436\\
        ET-Bert & 0.9944 & 0.9855 & \textbf{0.9788} & \underline{0.9456} & \textbf{0.9988} & \textbf{0.9987} & 0.9837 & 0.8151 & 0.9851 & 0.9842 & 0.9455 & 0.4296 & \underline{0.9325}\\
        \midrule
        \Lens~(Ours) & \textbf{0.9960} & \textbf{0.9898} & \underline{0.9752} & \textbf{0.9492} & 0.9951 & 0.9951 & \textbf{0.9963} & \textbf{0.9610} & \textbf{0.9877} & \textbf{0.9870} & \textbf{0.9878} & \textbf{0.6802} & \textbf{0.9633}\\
        \bottomrule
    \end{tabular}
    \end{adjustbox}
\end{table*}

In Tasks 6 and 8, \Lens~performs slightly better or comparably to baseline methods with both high accuracy and F1, demonstrating excellent generalization capabilities on an unseen dataset, even with 209 classes and 196 classes separately. In contrast, the TrafficLLM deteriorates as finetuning on limited data fails to mitigate the modality gap and learns the large label sets well. Additionally, TrafficLLM can not adapt well to Task 12 due to the skewed test label distribution. 

As shown in Table~\ref{tab:iot_f1_comparison}, we provide a detailed case study on Task 12 IoT method detection. \Lens~consistently outperforms the competitive baselines ET-BERT and YaTC across both head and long-tailed classes, with test sample sizes ranging from 5 to 11k. In particular, \Lens~accurately distinguishes DDoS from DoS, and effectively recognizes long-tailed classes such as Web-based and BruteForce, whereas other baselines malfunctioned on these classes. 

\begin{wraptable}[8]{r}{0.5\textwidth}
    \centering
    \vspace{-0.6in}
    \caption{The case study of classification F1 score in the Task 12 IoT attack method detection. \Lens~outperforms baselines well on both head and long-tailed classes. The n denotes the number of test samples. }
    \label{tab:iot_f1_comparison}
    \begin{adjustbox}{width=\linewidth}
    \begin{tabular}{lcccc}
        \toprule
        \textbf{Attack Type} & \textbf{Sample Size (n)} & \textbf{ET-BERT} & \textbf{YaTC} & \textbf{Lens (Ours)} \\
        \midrule
        DDoS        & 11055 & 0.973 & \underline{0.979} & \textbf{0.995} \\
        DoS         & 1216  & 0.746 & \underline{0.795} & \textbf{0.968} \\
        \hdashline
        Spoofing    & 55    & 0.486 & \underline{0.667} & \textbf{0.779} \\
        Web-based   & 10    & --    & --    & \textbf{0.308} \\
        BruteForce  & 5     & --    & \underline{0.500} & \textbf{0.571} \\
    \bottomrule
    \end{tabular}
    \end{adjustbox}
\end{wraptable}


\subsection{Network Traffic Extensibility Performance}\label{sec:extensibility}
We also evaluate \Lens's extensibility on Task 6 with 209 classes and Task 8 with 196 classes. Specifically, we set 3 scenarios where models need to face 1, 3, or 5 novel classes at each time, simulating the real network scenario, where models need to be updated frequently to classify new application types.

To evaluate the extensibility, we first finetune models on old classes until convergence. Then, we finetune only on data samples from new classes to test how well models extend the classification to new classes. Technically, we incorporate new-class options into \Lens’s finetuning context, whereas ET-BERT extends its MLP classifier by adding output dimensions for the new classes. Besides, we further incorporate the learning-without-forgetting (LwF) mechanism~\cite{li2017learning} into ET-BERT’s classifier for a more comprehensive comparison, denoted as ET-BERT-LwF. For both ET-BERT and ET-BERT-LwF, the newly added output dimensions of their MLP classifiers are initialized using Kaiming initialization~\cite{he2015delving}. All test scenarios include data from both old and new classes to enable fair performance comparison. Detailed experimental settings and results are provided in Appendix~\ref{appendix:extensibility}.

\begin{table*}[!h]
\centering
\small
\setlength{\tabcolsep}{3pt}
\caption{
    Extensibility performance on Task 6 (209 classes) and Task 8 (196 classes).
    \Lens~consistently achieves superior extension to unseen classes while maintaining strong accuracy on known classes.
}
\label{tab:extendibility}

\begin{adjustbox}{width=\textwidth}
\begin{tabular}{lcccccccccccc}
\toprule
\multirow{3}{*}{Scenarios} 
& \multicolumn{6}{c}{\textbf{Task 6 (209 classes)}} 
& \multicolumn{6}{c}{\textbf{Task 8 (196 classes)}} 
\\
\cmidrule(lr){2-7}
\cmidrule(lr){8-13}
& \multicolumn{2}{c}{1 new} 
& \multicolumn{2}{c}{3 new} 
& \multicolumn{2}{c}{5 new}
& \multicolumn{2}{c}{1 new} 
& \multicolumn{2}{c}{3 new} 
& \multicolumn{2}{c}{5 new} \\
\cmidrule(lr){2-7}
\cmidrule(lr){8-13}
& AC & F1 & AC & F1 & AC & F1 & AC & F1 & AC & F1 & AC & F1 \\
\midrule
ET-BERT 
& 0.8536 & 0.7832 & 0.6387 & 0.4400 & 0.7137 & 0.5906
& 0.8257 & 0.8026 & 0.8477 & 0.8695 & 0.6079 & 0.5762 \\
ET-BERT-LwF 
& \underline{0.9378} & \textbf{0.8688} 
& \underline{0.8783} & \underline{0.8625} 
& \underline{0.8261} & \textbf{0.8269}
& \underline{0.9211} & \textbf{0.8941}
& \underline{0.8416} & \textbf{0.9174}
& \underline{0.7862} & \textbf{0.8940} \\
\Lens (Ours)
& \textbf{0.9565} & \underline{0.8578}
& \textbf{0.9518} & \textbf{0.8659}
& \textbf{0.9199} & \underline{0.8264}
& \textbf{0.9397} & \underline{0.8861}
& \textbf{0.8962} & \underline{0.8801}
& \textbf{0.8730} & \underline{0.8407} \\
\bottomrule
\end{tabular}
\end{adjustbox}
\end{table*}
\textbf{As shown in Table~\ref{tab:extendibility}, we can observe \Lens~extends to new classes well with better accuracy and F1 compared to ET-BERT, while always achieves better accuracy than ET-BERT-LwF.} When extending to the challenging 5 new classes, \Lens~outperforms ET-BERT significantly with more than +20\% improvement of accuracy and F1 score on both tasks. Although ET-BERT-LwF performs much better than the original one, our \Lens~still outperforms with better accuracy. In the 1-new-class scenario, \Lens~achieves +10.85\% and +7.91\% average gains over ET-BERT in accuracy and F1, respectively, and consistently outperforms ET-BERT-LwF in accuracy. On Task 8 with 3 new classes, \Lens~achieves higher accuracy and comparable F1 scores compared to both ET-BERT and ET-BERT-LwF.

Nevertheless, on Task 6 with three new classes, ET-BERT reallocates logits aggressively toward the new classes, resulting in 31.31\% and 42.59\% drops in accuracy and F1 compared with \Lens. In contrast, our reformulation enables \Lens~to extend to classify new classes more seamlessly via context-aware finetuning. Although ET-BERT-LwF alleviates the drastic shift toward new classes, it still struggles to balance old and new classes and yields lower overall accuracy.

In this paper, we aim to show our model's extensibility to new classes rather than developing new continual learning methods. Although continual learning could further enhance \Lens~'s capability, we leave this for future work. More detailed performance on each class are listed in Table~\ref{tab:1novelclass}, \ref{tab:3novelclass}, and \ref{tab:5novelclass} of Appendix~\ref{appendix:extensibility}.


\subsection{Network Traffic Generation Performance} \label{sec:generation}
In addition, we evaluate \Lens's performance on traffic generation tasks and compare with SOTA baselines, including NetShare~\cite{netshare}, and TrafficLLM~\cite{cui2025trafficllm}. Based on the methodology of NetShare~\cite{netshare}, we generate five vital network header fields: source IP, destination IP, source port, destination port, and packet length at the packet level. The generation of these synthetic header fields can help create usable pcap traffic for network simulation. For each dataset in our experiments, all generation tasks are individually conducted using a supervised finetuning approach. For the metrics JSD and TVD, lower values indicate a smaller distribution difference between the generated and ground-truth data. 

\begin{table*}[!thb]
    \centering
    \small
    \caption{Performance on traffic generation tasks in terms of JSD and TVD. We can see that \Lens~outperforms baselines consistently on all datasets in generating Destination Port. In source IP generation, \Lens ~achieves comparable or better performances than baselines. Bold numbers denote the best results, while underlined numbers are the second best. }
    \label{tab:compGenCombined}
    \begin{adjustbox}{width=\textwidth}
    \begin{tabular}{llcccccccccc}
        \toprule
         \multirow{2}{*}{\centering Datasets} & \multirow{2}{*}{\centering Method} & \multicolumn{5}{c}{JSD $\downarrow$ } & \multicolumn{5}{c}{TVD $\downarrow$ } \\
        \cmidrule(lr){3-7}
        \cmidrule(lr){8-12}
         & & Src IP & Dst IP & Src Port & Dst Port & Len & Src IP & Dst IP & Src Port & Dst Port & Len \\
         \midrule
         \multirow{3}{*}{ISCX-VPN}  & NetShare & 0.3591 & 0.3787 & 0.6539 & 0.5893 & 0.6793 
                                               & 0.5802 & 0.5948 & 0.9632 & 0.9137 & 0.9893 \\
                                    & TrafficLLM & \textbf{0.0946} & \underline{0.1175} & \underline{0.5742} & \underline{0.0430} & \underline{0.0513} 
                                                 & \underline{0.1849} & \underline{0.1772} & \underline{0.5920} & \underline{0.0600} & \underline{0.0845} \\
                                    & \Lens~(Ours) & \underline{0.0974} & \textbf{0.0905} & \textbf{0.5574} & \textbf{0.0271} & \textbf{0.0338} 
                                                   & \textbf{0.1719} & \textbf{0.1245} & \textbf{0.5789} & \textbf{0.0343} & \textbf{0.0469}\\
        \midrule
        \multirow{3}{*}{ISCXTor}    & NetShare & 0.3084 & \underline{0.4160} & 0.5835 & 0.5736 & 0.6531 
                                               & 0.4930 & 0.6436 & 0.8813 & 0.8807 & 0.9756 \\
                                    & TrafficLLM & \underline{0.0023} & \textbf{0.3629} & \textbf{0.5635} & \underline{0.1770} & \textbf{0.0359} 
                                                 & \underline{0.0047} & \textbf{0.4519} & \textbf{0.5838} & \underline{0.2339} & \textbf{0.0500} \\
                                    & \Lens~(Ours) & \textbf{0.0022} & 0.4842 & \underline{0.5826} & \textbf{0.1337} & \underline{0.0398} 
                                                   & \textbf{0.0038} & \underline{0.5620} & \underline{0.6133} & \textbf{0.1877} & \underline{0.0560} \\
        \midrule
        \multirow{3}{*}{USTC-TFC}   & NetShare & 0.4415 & 0.5065 & 0.5731 & 0.5885 & 0.6826 
                                               & 0.6876 & 0.7794 & 0.9037 & 0.9114 & 0.9933 \\
                                    & TrafficLLM & \textbf{0.3702} & \textbf{0.4186} & \textbf{0.3496} & \underline{0.2746} & \underline{0.0159} 
                                                 & \underline{0.3915} & \textbf{0.4731} & \textbf{0.3656} & \underline{0.3261} & \underline{0.0259}\\
                                    & \Lens~(Ours) & \underline{0.3783} & \underline{0.4361} & \underline{0.3864} & \textbf{0.2685} & \textbf{0.0143} 
                                                   & \textbf{0.3910} & \underline{0.4748} & \underline{0.4076} & \textbf{0.2901} & \textbf{0.0203} \\
        \midrule
        \multirow{3}{*}{Cross Platform (IOS)} & NetShare & 0.3304 & 0.5267 & \underline{0.6056} & 0.6104 & 0.6120 
                                                         & 0.3389 & 0.5188 & 0.9289 & 0.9318 & 0.9491 \\
                                    & TrafficLLM & \textbf{0.0003} & \textbf{0.2433} & \textbf{0.5784} & \underline{0.0134} & \textbf{0.0521} 
                                                 & \textbf{0.0006} & \textbf{0.3374} & \textbf{0.6006} & \underline{0.0265} & \textbf{0.0662}\\
                                    & \Lens~(Ours) & \textbf{0.0003} & \underline{0.3241} & 0.6508 & \textbf{0.0083} & \underline{0.0608} 
                                                   & \textbf{0.0006} & \underline{0.4523} & \underline{0.6746} & \textbf{0.0132} & \underline{0.0780}\\
        \midrule
        \multirow{3}{*}{Cross Platform (AN)} & NetShare & 0.3459 & 0.3964 & \underline{0.6219} & 0.6269 & 0.6312 
                                                        & 0.5437 & 0.6205 & 0.9353 & 0.9443 & 0.9598\\
                                    & TrafficLLM & \underline{0.0016} & \textbf{0.1969} & \textbf{0.6011} & \underline{0.0082} & \textbf{0.0672} 
                                                 & \underline{0.0216} & \textbf{0.3101} & \textbf{0.6220} & \underline{0.0377} & \textbf{0.0835}\\
                                    & \Lens~(Ours) & \textbf{0.0003} & \underline{0.2809} & 0.6531 & \textbf{0.0046} & \underline{0.0690} 
                                                   & \textbf{0.0041} & \underline{0.4166} & \underline{0.6796} & \textbf{0.0130} & \underline{0.0872} \\
        \midrule
        \multirow{3}{*}{CIRA-CIC-DoHBrw} & NetShare & 0.3955 & 0.6117 & \underline{0.5446} & 0.4630 & 0.6566 
                                                    & 0.3886 & 0.6315 & 0.8825 & 0.7754 & 0.9776\\
                                         & TrafficLLM & \underline{0.0162} & \textbf{0.3782} & \textbf{0.4858} & \textbf{0.0001} & \underline{0.0614} 
                                                      & \underline{0.0838} & \textbf{0.4676} & \textbf{0.5258} & \textbf{0.0003} & \underline{0.1069} \\
                                         & \Lens~(Ours) & \textbf{0.0041} & \underline{0.4105} & 0.6915 & \textbf{0.0001} & \textbf{0.0481} 
                                                      & \textbf{0.0246} & \underline{0.4896} & \underline{0.7065} & \textbf{0.0003} & \textbf{0.0728}\\
        \midrule
        \multirow{3}{*}{CIC-IoT-2023} & NetShare & 0.0732 & 0.0804 & 0.5939 & 0.1188 & 0.6807
                                                 & 0.2155& 0.2273 & 0.8920 & 0.2936 & 0.9904 \\
                                              & TrafficLLM & \underline{0.0598} & \underline{0.0345} & \underline{0.5544} & \underline{0.0586} & \textbf{0.0039} 
                                                           & \underline{0.1523} & \underline{0.1072} & \underline{0.5844} & \underline{0.0833} & \underline{0.0061}\\
                                              & \Lens~(Ours) & \textbf{0.0146} & \textbf{0.0098} & \textbf{0.0179} & \textbf{0.0262} & \textbf{0.0039} & \textbf{0.0779} & \textbf{0.0325} & \textbf{0.0217} & \textbf{0.0295} & \textbf{0.0058}\\
         \bottomrule
    \end{tabular}
    \end{adjustbox}
\end{table*}

From the Table~\ref{tab:compGenCombined}, on the generation of destination port, \Lens~outperforms Netshare and TrafficLLM consistently over the 6 datasets. \textbf{\Lens~reduces JSD and TVD up to 55.29\% and 64.59\%, indicating \Lens's significantly better understanding of destination port numbers.} Notably, even \Lens~does not pre-train on the CrossPlatform dataset, it still surpasses the TrafficLLM by +38.06\% and +50.19\% on JSD and TVD, respectively. This is because our KG-MSP intentionally learns the networking knowledge of port numbers, correlating specific destination port numbers with typical protocol types. 

Besides, \Lens~generates more aligned source IP with lower TVD on all datasets, which benefits from the auxiliary context pretraining that correlates dataset sources with network traffic. For source port and destination IP generation, TrafficLLM performs slightly better than \Lens. This is because the TrafficLLM has more parameters that can memorize randomly assigned source port numbers and varied destination IP addresses better. However, even TrafficLLM has way more parameters, \Lens~excels over TrafficLLM in most packet length generation tasks. Lastly, pretraining-based \Lens~and TrafficLLM perform better than GAN-based Netshare on all tasks. This is because pretraining-based models learn better packet-level network representation, while Netshare sacrifices it for global distribution similarity.


\noindent\textbf{Network Traffic Generation Fuzzing Tests.}
To evaluate the fidelity of the generated network traffic, we conduct network fuzzing tests on the IoT Malicious detection task (Task 11). First, we resume \Lens~from its finetuned checkpoint and generate network header fields using the Task 11 training data. Then, machine learning models such as Decision Tree, SVM and MLP, are trained on the generated network header fields as a binary classifier to detect malicious network traffic. Lastly, the trained machine learning models are evaluated on the real Task 11 test set for malicious traffic detection, thereby assessing the quality of the synthetic network data.

\begin{figure*}[!th]
    \centering
    \includegraphics[width=0.8\textwidth]{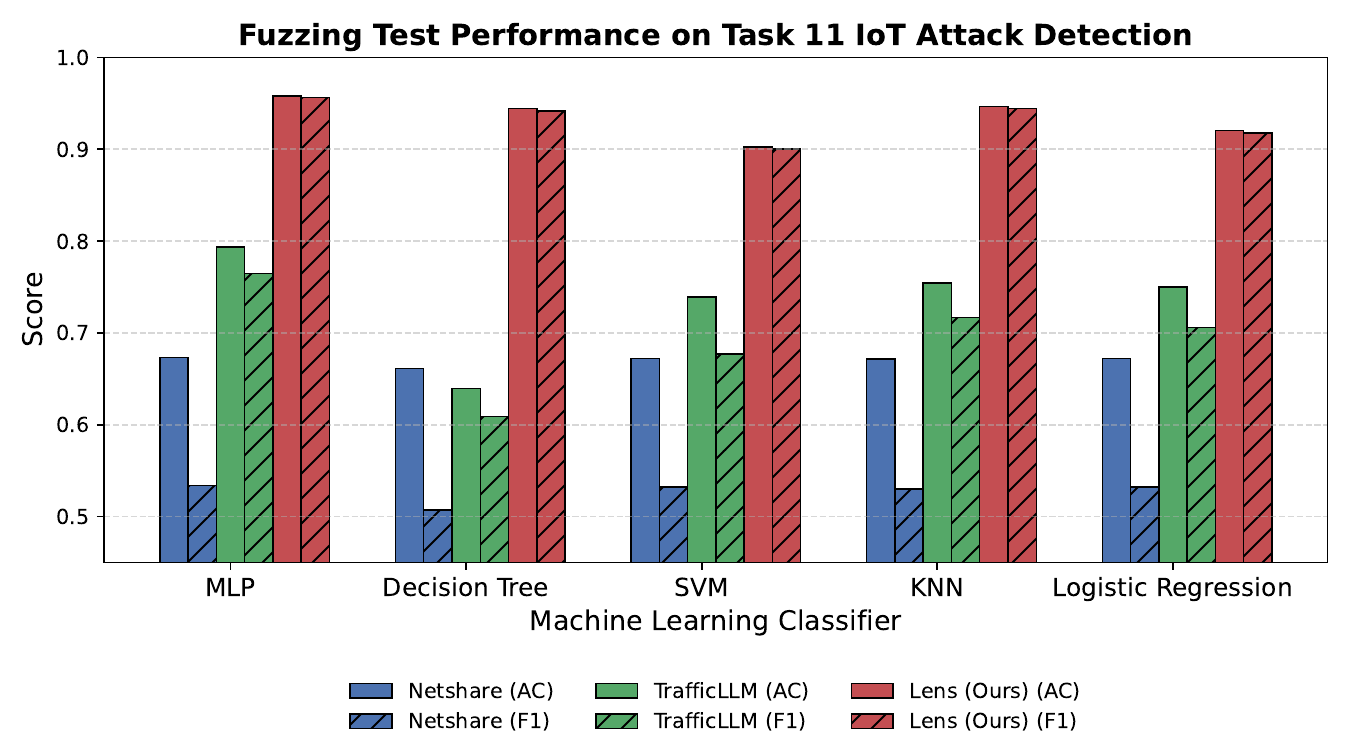}
    \caption{Fuzzing performance on IoT attack detection. Machine-learning models trained on \Lens-generated traffic achieve consistently higher accuracy and F1 than those trained on baselines’ generated traffic.}
    \label{fig:fuzzing}
\end{figure*}

From Figure~\ref{fig:fuzzing}, we can observe \textbf{\Lens~generates high-fidelity network traffic that aligns better with real scenarios and achieves substantially better performance than baselines on all machine learning-based detectors.} Specifically, \Lens~performs better than the second best TrafficLLM with +30.46\% and +33.3\% improvement on accuracy and F1. As the payload is usually encrypted, users can either replay saved network payload or generate them randomly with an encrypted algorithm like AES~\cite{brown2023advanced} for a more practical use. 


\subsection{Ablation Studies} \label{sec:ablation}

\textbf{KG-MSP pretraining and context-aware finetuning contribute to classification and generation.}



\begin{wraptable}[11]{r}{0.4\textwidth}
    \centering
    \small
    \vspace{-0.15in}
    \caption{Ablation studies on VPN
Destination Port generation. The pretraining KG-MSP and finetuning context improves both JSD and TVD.}
    \label{tab:gen}
    \begin{adjustbox}{width=0.95\linewidth}
    \begin{tabular}{l c c }
        \toprule
        Settings & JSD $\downarrow$  & TVD  $\downarrow$ \\
        \midrule
        \texttt{\Lens~(Full model)} & \textbf{0.0271} & \textbf{0.0343} \\
        \midrule
        \hspace{5mm}\texttt{w/o FT-Context} & 0.0294 & 0.0384\\
        \hspace{5mm}\texttt{w/o KG-MSP} & 0.0308 & 0.0437 \\
        \bottomrule
    \end{tabular}
    \end{adjustbox}
\end{wraptable}
For network traffic classification, we conduct ablation studies on Tor service detection (Task 4) and VPN application classification (Task 3). In Table~\ref{tab:ablation-both}, both KG-MSP and context-aware finetuning (FT-Context) yield substantial gains, improving F1 by +4.2\% and +4.99\% on Tor service detection. For VPN application classification, KG-MSP boosts accuracy and F1 by +11.6\% and +5.2\%, and FT-Context provides an additional +5.25\% accuracy improvement. For network traffic generation, we evaluate the VPN destination-port task. As shown in Table~\ref{tab:gen}, FT-Context reduces JSD and TVD by 7.82\% and 10.68\%, while KG-MSP provides further reductions of 12.8\% and 21.5\%. Overall, KG-MSP strengthens representation learning by masking key metadata and payload information, while FT-Context reduces the modality gap during finetuning. These results demonstrate that both KG-MSP and FT-Context are essential for both network traffic classification and generation.

\begin{table*}[!thb]
\centering
\small
\setlength{\tabcolsep}{3pt}
\caption{
Ablation studies on Tor service detection and VPN application classification.
Both KG-MSP and FT-Context contribute significantly to performance improvement.
}
\label{tab:ablation-both}
\begin{tabular}{lcccccc}
\toprule
\multirow{2}{*}{Settings}
& \multicolumn{2}{c}{\textbf{Tor Service Detection}}
& \multicolumn{2}{c}{\textbf{VPN App. Classification}} \\
\cmidrule(lr){2-3}
\cmidrule(lr){4-5}
& AC & F1 & AC & F1 \\
\midrule
\texttt{\Lens~(Full model)} 
& \textbf{0.9692} & \textbf{0.8120}
& \textbf{0.8406} & \textbf{0.8137} \\
\midrule
\hspace{3mm}\texttt{w/o FT-Context}
& 0.9577 & 0.7700
& 0.7881 & 0.8111 \\
\hspace{3mm}\texttt{w/o KG-MSP}
& 0.9612 & 0.7621
& 0.7246 & 0.7622 \\
\bottomrule
\end{tabular}
\end{table*}

\section{Discussion and Conclusion}
In this paper, we proposed \Lens, a unified knowledge-guided foundation model for network traffic excelling in both network traffic classification and generation. Through pretraining with Knowledge-guided Mask Span Prediction with textual context, \Lens~effectively learns generalizable network representations from large-scale unlabeled traffic data. To effectively extend classification to new classes, we reframe traffic classification as a closed-ended generation task and handle the distribution shifts of new-class data through context‐aware finetuning. Finally, we evaluated the performance of \Lens~on 6 real-world datasets, including 12 traffic classification tasks and 5 network generation tasks. For traffic classification, extensive experimental results demonstrated that \Lens~outperforms the baselines with a large margin in most tasks, showcasing its strong effectiveness and extensibility. For traffic generation, \Lens~generates better high-fidelity network traffic for effective network simulations, outperforming baselines substantially in fuzzing tests. Furthermore, ablation studies validate the effectiveness of KG-MSP and context-aware finetuning. In the future, we plan to implement the proposed method in real-world computer systems to evaluate its performance.



\newpage
}

\bibliographystyle{abbrv}
\bibliography{reference.bib}

@String{Computing = "Computing" }

@String{Computer = "{IEEE} Computer" }

@String{Springer = "Springer-Verlag" }

@inproceedings{cumul,
  title={Website Fingerprinting at Internet Scale.},
  author={Panchenko, Andriy and Lanze, Fabian and Pennekamp, Jan and Engel, Thomas and Zinnen, Andreas and Henze, Martin and Wehrle, Klaus},
  booktitle={NDSS},
  year={2016}
}

@article{appscanner,
  title={Robust smartphone app identification via encrypted network traffic analysis},
  author={Taylor, Vincent F and Spolaor, Riccardo and Conti, Mauro and Martinovic, Ivan},
  journal={IEEE Transactions on Information Forensics and Security},
  volume={13},
  number={1},
  pages={63--78},
  year={2017},
  publisher={IEEE}
}

@inproceedings{bind,
  title={Adaptive encrypted traffic fingerprinting with bi-directional dependence},
  author={Al-Naami, Khaled and Chandra, Swarup and Mustafa, Ahmad and Khan, Latifur and Lin, Zhiqiang and Hamlen, Kevin and Thuraisingham, Bhavani},
  booktitle={Proceedings of the 32nd Annual Conference on Computer Security Applications},
  pages={177--188},
  year={2016}
}

@inproceedings{isanon,
  title={isAnon: Flow-based anonymity network traffic identification using extreme gradient boosting},
  author={Cai, Zhenzhen and Jiang, Bo and Lu, Zhigang and Liu, Junrong and Ma, Pingchuan},
  booktitle={2019 International Joint Conference on Neural Networks (IJCNN)},
  pages={1--8},
  year={2019},
  organization={IEEE}
}

@inproceedings{df,
  title={Deep fingerprinting: Undermining website fingerprinting defenses with deep learning},
  author={Sirinam, Payap and Imani, Mohsen and Juarez, Marc and Wright, Matthew},
  booktitle={Proceedings of the 2018 ACM SIGSAC Conference on Computer and Communications Security},
  pages={1928--1943},
  year={2018}
}

@inproceedings{fs-net,
  title={Fs-net: A flow sequence network for encrypted traffic classification},
  author={Liu, Chang and He, Longtao and Xiong, Gang and Cao, Zigang and Li, Zhen},
  booktitle={IEEE INFOCOM 2019-IEEE Conference On Computer Communications},
  pages={1171--1179},
  year={2019},
  organization={IEEE}
}

@article{deeppacket,
  title={Deep packet: A novel approach for encrypted traffic classification using deep learning},
  author={Lotfollahi, Mohammad and Jafari Siavoshani, Mahdi and Shirali Hossein Zade, Ramin and Saberian, Mohammdsadegh},
  journal={Soft Computing},
  volume={24},
  number={3},
  pages={1999--2012},
  year={2020},
  publisher={Springer}
}

@article{tscrnn,
  title={TSCRNN: A novel classification scheme of encrypted traffic based on flow spatiotemporal features for efficient management of IIoT},
  author={Lin, Kunda and Xu, Xiaolong and Gao, Honghao},
  journal={Computer Networks},
  volume={190},
  pages={107974},
  year={2021},
  publisher={Elsevier}
}

@article{bilstm,
  title={Identification of encrypted traffic through attention mechanism based long short term memory},
  author={Yao, Haipeng and Liu, Chong and Zhang, Peiying and Wu, Sheng and Jiang, Chunxiao and Yu, Shui},
  journal={IEEE Transactions on Big Data},
  volume={8},
  number={1},
  pages={241--252},
  year={2019},
  publisher={IEEE}
}

@article{datanet,
  title={Datanet: Deep learning based encrypted network traffic classification in sdn home gateway},
  author={Wang, Pan and Ye, Feng and Chen, Xuejiao and Qian, Yi},
  journal={IEEE Access},
  volume={6},
  pages={55380--55391},
  year={2018},
  publisher={IEEE}
}

@inproceedings{pert,
  title={PERT: Payload encoding representation from transformer for encrypted traffic classification},
  author={He, Hong Ye and Yang, Zhi Guo and Chen, Xiang Ning},
  booktitle={2020 ITU Kaleidoscope: Industry-Driven Digital Transformation (ITU K)},
  pages={1--8},
  year={2020},
  organization={IEEE}
}

@inproceedings{etbert,
  title={Et-bert: A contextualized datagram representation with pre-training transformers for encrypted traffic classification},
  author={Lin, Xinjie and Xiong, Gang and Gou, Gaopeng and Li, Zhen and Shi, Junzheng and Yu, Jing},
  booktitle={Proceedings of the ACM Web Conference 2022},
  pages={633--642},
  year={2022}
}

@inproceedings{yatc,
  title={Yet another traffic classifier: a masked autoencoder based traffic transformer with multi-level flow representation},
  author={Zhao, Ruijie and Zhan, Mingwei and Deng, Xianwen and Wang, Yanhao and Wang, Yijun and Gui, Guan and Xue, Zhi},
  booktitle={Proceedings of the AAAI Conference on Artificial Intelligence},
  volume={37},
  number={4},
  pages={5420--5427},
  year={2023}
}

@article{netgpt,
  title={NetGPT: Generative Pretrained Transformer for Network Traffic},
  author={Meng, Xuying and Lin, Chungang and Wang, Yequan and Zhang, Yujun},
  journal={arXiv preprint arXiv:2304.09513},
  year={2023}
}

@article{ns-3,
  title={Network simulations with the ns-3 simulator},
  author={Henderson, Thomas R and Lacage, Mathieu and Riley, George F and Dowell, Craig and Kopena, Joseph},
  journal={SIGCOMM demonstration},
  volume={14},
  number={14},
  pages={527},
  year={2008}
}

@inproceedings{yans,
  title={Yet another network simulator},
  author={Lacage, Mathieu and Henderson, Thomas R},
  booktitle={Proceedings of the 2006 Workshop on ns-3},
  pages={12--es},
  year={2006}
}

@inproceedings{dynamo,
  title={Generating representative, live network traffic out of millions of code repositories},
  author={B{\"u}hler, Tobias and Schmid, Roland and Lutz, Sandro and Vanbever, Laurent},
  booktitle={Proceedings of the 21st ACM Workshop on Hot Topics in Networks},
  pages={1--7},
  year={2022}
}

@article{botta2gentool,
  title={A tool for the generation of realistic network workload for emerging networking scenarios},
  author={Botta, Alessio and Dainotti, Alberto and Pescap{\'e}, Antonio},
  journal={Computer Networks},
  volume={56},
  number={15},
  pages={3531--3547},
  year={2012},
  publisher={Elsevier}
}

@article{harpoon,
  title={Harpoon: a flow-level traffic generator for router and network tests},
  author={Sommers, Joel and Kim, Hyungsuk and Barford, Paul},
  journal={ACM SIGMETRICS Performance Evaluation Review},
  volume={32},
  number={1},
  pages={392--392},
  year={2004},
  publisher={ACM New York, NY, USA}
}

@article{swing,
  title={Swing: Realistic and responsive network traffic generation},
  author={Vishwanath, Kashi Venkatesh and Vahdat, Amin},
  journal={IEEE/ACM Transactions on Networking},
  volume={17},
  number={3},
  pages={712--725},
  year={2009},
  publisher={IEEE}
}

@inproceedings{DoppelGANger,
  title={Using gans for sharing networked time series data: Challenges, initial promise, and open questions},
  author={Lin, Zinan and Jain, Alankar and Wang, Chen and Fanti, Giulia and Sekar, Vyas},
  booktitle={Proceedings of the ACM Internet Measurement Conference},
  pages={464--483},
  year={2020}
}

@inproceedings{netshare,
  title={Practical gan-based synthetic ip header trace generation using netshare},
  author={Yin, Yucheng and Lin, Zinan and Jin, Minhao and Fanti, Giulia and Sekar, Vyas},
  booktitle={Proceedings of the ACM SIGCOMM 2022 Conference},
  pages={458--472},
  year={2022}
}

@inproceedings{gan2022,
  title={Knowledge enhanced gan for IoT traffic generation},
  author={Hui, Shuodi and Wang, Huandong and Wang, Zhenhua and Yang, Xinghao and Liu, Zhongjin and Jin, Depeng and Li, Yong},
  booktitle={Proceedings of the ACM Web Conference 2022},
  pages={3336--3346},
  year={2022}
}

@article{gan2021,
  title={Synthetic flow-based cryptomining attack generation through Generative Adversarial Networks},
  author={Mozo, Alberto and Gonz{\'a}lez-Prieto, {\'A}ngel and Pastor, Antonio and G{\'o}mez-Canaval, Sandra and Talavera, Edgar},
  journal={Scientific reports},
  volume={12},
  number={1},
  pages={2091},
  year={2022},
  publisher={Nature Publishing Group UK London}
}

@article{first-gan,
  title={Flow-based network traffic generation using generative adversarial networks},
  author={Ring, Markus and Schl{\"o}r, Daniel and Landes, Dieter and Hotho, Andreas},
  journal={Computers \& Security},
  volume={82},
  pages={156--172},
  year={2019},
  publisher={Elsevier}
}

@article{t5,
  title={Exploring the limits of transfer learning with a unified text-to-text transformer},
  author={Raffel, Colin and Shazeer, Noam and Roberts, Adam and Lee, Katherine and Narang, Sharan and Matena, Michael and Zhou, Yanqi and Li, Wei and Liu, Peter J},
  journal={The Journal of Machine Learning Research},
  volume={21},
  number={1},
  pages={5485--5551},
  year={2020},
  publisher={JMLRORG}
}

@article{codet5,
  title={Codet5: Identifier-aware unified pre-trained encoder-decoder models for code understanding and generation},
  author={Wang, Yue and Wang, Weishi and Joty, Shafiq and Hoi, Steven CH},
  journal={arXiv preprint arXiv:2109.00859},
  year={2021}
}

@article{bert,
  title={Bert: Pre-training of deep bidirectional transformers for language understanding},
  author={Devlin, Jacob and Chang, Ming-Wei and Lee, Kenton and Toutanova, Kristina},
  journal={arXiv preprint arXiv:1810.04805},
  year={2018}
}

@article{knn,
  title={$ k $-Nearest Neighbor Augmented Neural Networks for Text Classification},
  author={Wang, Zhiguo and Hamza, Wael and Song, Linfeng},
  journal={arXiv preprint arXiv:1708.07863},
  year={2017}
}

@article{albert,
  title={Albert: A lite bert for self-supervised learning of language representations},
  author={Lan, Zhenzhong and Chen, Mingda and Goodman, Sebastian and Gimpel, Kevin and Sharma, Piyush and Soricut, Radu},
  journal={arXiv preprint arXiv:1909.11942},
  year={2019}
}

@article{gan,
  title={Generative adversarial nets},
  author={Goodfellow, Ian and Pouget-Abadie, Jean and Mirza, Mehdi and Xu, Bing and Warde-Farley, David and Ozair, Sherjil and Courville, Aaron and Bengio, Yoshua},
  journal={Advances in neural information processing systems},
  volume={27},
  year={2014}
}

@inproceedings{vpn,
  title={Characterization of encrypted and vpn traffic using time-related},
  author={Draper-Gil, Gerard and Lashkari, Arash Habibi and Mamun, Mohammad Saiful Islam and Ghorbani, Ali A},
  booktitle={Proceedings of the 2nd international conference on information systems security and privacy (ICISSP)},
  pages={407--414},
  year={2016}
}

@conference{tor,
author={Arash {Habibi Lashkari}. and Gerard {Draper Gil}. and Mohammad Saiful Islam Mamun. and Ali A. Ghorbani.},
title={Characterization of Tor Traffic using Time based Features},
booktitle={Proceedings of the 3rd International Conference on Information Systems Security and Privacy - ICISSP},
year={2017},
pages={253-262},
publisher={SciTePress},
organization={INSTICC},
doi={10.5220/0006105602530262},
isbn={978-989-758-209-7},
issn={2184-4356},
}

@inproceedings{doh,
  title={Detection of doh tunnels using time-series classification of encrypted traffic},
  author={MontazeriShatoori, Mohammadreza and Davidson, Logan and Kaur, Gurdip and Lashkari, Arash Habibi},
  booktitle={2020 IEEE Intl Conf on Dependable, Autonomic and Secure Computing, Intl Conf on Pervasive Intelligence and Computing, Intl Conf on Cloud and Big Data Computing, Intl Conf on Cyber Science and Technology Congress (DASC/PiCom/CBDCom/CyberSciTech)},
  pages={63--70},
  year={2020},
  organization={IEEE}
}

@article{guthula2023netfound,
  title={netFound: Foundation model for network security},
  author={Guthula, Satyandra and Beltiukov, Roman and Battula, Navya and Guo, Wenbo and Gupta, Arpit},
  journal={arXiv preprint arXiv:2310.17025},
  year={2023}
}

@article{iot,
  title={CICIoT2023: A real-time dataset and benchmark for large-scale attacks in IoT environment},
  author={Neto, Euclides Carlos Pinto and Dadkhah, Sajjad and Ferreira, Raphael and Zohourian, Alireza and Lu, Rongxing and Ghorbani, Ali A},
  year={2023},
  publisher={Preprints}
}

@inproceedings{ustc,
  title={Malware traffic classification using convolutional neural network for representation learning},
  author={Wang, Wei and Zhu, Ming and Zeng, Xuewen and Ye, Xiaozhou and Sheng, Yiqiang},
  booktitle={2017 International conference on information networking (ICOIN)},
  pages={712--717},
  year={2017},
  organization={IEEE}
}

@inproceedings{cp,
  title={Flowprint: Semi-supervised mobile-app fingerprinting on encrypted network traffic},
  author={Van Ede, Thijs and Bortolameotti, Riccardo and Continella, Andrea and Ren, Jingjing and Dubois, Daniel J and Lindorfer, Martina and Choffnes, David and van Steen, Maarten and Peter, Andreas},
  booktitle={Network and distributed system security symposium (NDSS)},
  volume={27},
  year={2020}
}

@article{adamw,
  title={Decoupled weight decay regularization},
  author={Loshchilov, Ilya and Hutter, Frank},
  journal={arXiv preprint arXiv:1711.05101},
  year={2017}
}

@article{oliveira2016computer,
  title={Computer network traffic prediction: a comparison between traditional and deep learning neural networks},
  author={Oliveira, Tiago Prado and Barbar, Jamil Salem and Soares, Alexsandro Santos},
  journal={International Journal of Big Data Intelligence},
  volume={3},
  number={1},
  pages={28--37},
  year={2016},
  publisher={Inderscience Publishers (IEL)}
}

@article{qian2024netbench,
  title={NetBench: A Large-Scale and Comprehensive Network Traffic Benchmark Dataset for Foundation Models},
  author={Qian, Chen and Li, Xiaochang and Wang, Qineng and Zhou, Gang and Shao, Huajie},
  journal={arXiv preprint arXiv:2403.10319},
  year={2024}
}

@article{cui2025trafficllm,
  title={Trafficllm: Enhancing large language models for network traffic analysis with generic traffic representation},
  author={Cui, Tianyu and Lin, Xinjie and Li, Sijia and Chen, Miao and Yin, Qilei and Li, Qi and Xu, Ke},
  journal={arXiv preprint arXiv:2504.04222},
  year={2025}
}

@inproceedings{bbpe,
  title={Neural machine translation with byte-level subwords},
  author={Wang, Changhan and Cho, Kyunghyun and Gu, Jiatao},
  booktitle={Proceedings of the AAAI conference on artificial intelligence},
  volume={34},
  number={05},
  pages={9154--9160},
  year={2020}
}

@article{li2017learning,
  title={Learning without forgetting},
  author={Li, Zhizhong and Hoiem, Derek},
  journal={IEEE transactions on pattern analysis and machine intelligence},
  volume={40},
  number={12},
  pages={2935--2947},
  year={2017},
  publisher={IEEE}
}

@article{jiang2024netdiffusion,
  title={NetDiffusion: Network Data Augmentation Through Protocol-Constrained Traffic Generation},
  author={Jiang, Xi and Liu, Shinan and Gember-Jacobson, Aaron and Bhagoji, Arjun Nitin and Schmitt, Paul and Bronzino, Francesco and Feamster, Nick},
  journal={Proceedings of the ACM on Measurement and Analysis of Computing Systems},
  volume={8},
  number={1},
  pages={1--32},
  year={2024},
  publisher={ACM New York, NY, USA}
}

@inproceedings{decao2021autoregressive,
  author    = {Nicola {De Cao} and
               Gautier Izacard and
               Sebastian Riedel and
               Fabio Petroni},
  title     = {Autoregressive Entity Retrieval},
  booktitle = {9th International Conference on Learning Representations, {ICLR} 2021,
               Virtual Event, Austria, May 3-7, 2021},
  year      = {2021},
}

@article{zhao2025language,
  title={Language of Network: A Generative Pre-trained Model for Encrypted Traffic Comprehension},
  author={Zhao, Di and Jiang, Bo and Liu, Song and Cui, Susu and Shen, Meng and Han, Dongqi and Guan, Xingmao and Lu, Zhigang},
  journal={arXiv preprint arXiv:2505.19482},
  year={2025}
}

@article{qu2024trafficgpt,
  title={Trafficgpt: Breaking the token barrier for efficient long traffic analysis and generation},
  author={Qu, Jian and Ma, Xiaobo and Li, Jianfeng},
  journal={arXiv preprint arXiv:2403.05822},
  year={2024}
}

@inproceedings{wang2024netmamba,
  title={Netmamba: Efficient network traffic classification via pre-training unidirectional mamba},
  author={Wang, Tongze and Xie, Xiaohui and Wang, Wenduo and Wang, Chuyi and Zhao, Youjian and Cui, Yong},
  booktitle={2024 IEEE 32nd International Conference on Network Protocols (ICNP)},
  pages={1--11},
  year={2024},
  organization={IEEE}
}

@inproceedings{chu2024feasibility,
  title={Feasibility of state space models for network traffic generation},
  author={Chu, Andrew and Jiang, Xi and Liu, Shinan and Bhagoji, Arjun and Bronzino, Francesco and Schmitt, Paul and Feamster, Nick},
  booktitle={Proceedings of the 2024 SIGCOMM Workshop on Networks for AI Computing},
  pages={9--17},
  year={2024}
}

@software{Knowledgator_TurboT5_2023,
      author       = {{Knowledgator}},
      title        = {TurboT5},
      url          = {https://github.com/Knowledgator/TurboT5},
      year         = {2023}
}

@software{wireshark,
  author       = {Combs, Gerald and The Wireshark team},
  title        = {Wireshark},
  version      = {4.4.8},
  year         = {2024},
  organization = {The Wireshark Foundation},
  url          = {https://www.wireshark.org},
  urldate      = {2025-07-18},
}

@misc{miranda2021ultimateutils,
    title={Ultimate Utils - the Ultimate Utils Library for Machine Learning and Artificial Intelligence},
    author={Brando Miranda},
    year={2021},
    url={https://github.com/brando90/ultimate-utils},
    note={Available at: \url{https://www.ideals.illinois.edu/handle/2142/112797}}
}

@inproceedings{he2015delving,
  title={Delving deep into rectifiers: Surpassing human-level performance on imagenet classification},
  author={He, Kaiming and Zhang, Xiangyu and Ren, Shaoqing and Sun, Jian},
  booktitle={Proceedings of the IEEE international conference on computer vision},
  pages={1026--1034},
  year={2015}
}

@article{munea2016network,
  title={Network protocol fuzz testing for information systems and applications: a survey and taxonomy},
  author={Munea, Tewodros Legesse and Lim, Hyunwoo and Shon, Taeshik},
  journal={Multimedia tools and applications},
  volume={75},
  number={22},
  pages={14745--14757},
  year={2016},
  publisher={Springer}
}

@article{brown2023advanced,
  title={Advanced Encryption Standard (AES)},
  author={Brown, Karen H},
  journal={National Institute of Standards and Technology, Federal Information Processing Standards Publication (FIPS 197), US Department of Commerce, Updated: May},
  volume={9},
  year={2023}
}

@misc{splitcap,
  author = {{Netresec}},
  title = {SplitCap},
  howpublished = {\url{https://www.netresec.com/?page=SplitCap}},
  year = {2025}
}

@misc{tshark,
  author = {{Wireshark Foundation}},
  title = {TShark - The Wireshark Network Analyzer},
  howpublished = {\url{https://www.wireshark.org/docs/man-pages/tshark.html}},
  year = {2025}
}

\clearpage
\appendix
\section{More Details on Datasets and Preprocessing}
\subsection{Preprocessing Pipeline}\label{appendix:datapreprocess}
Firstly, we segment raw traffic into flows using SplitCap~\cite{splitcap} 
based on the standard 5-tuple (source/destination IP, ports, and protocol). Next, we parse each flow with Tshark~\cite{tshark} to extract the network, 
transport, and application-layer metadata, and append a 12-byte hexadecimal 
application payload following NetFound~\cite{guthula2023netfound}. 
Finally, we anonymize all source and destination IP addresses using two 
special placeholder tokens, consistent with YaTC~\cite{yatc}, to ensure 
privacy and prevent identifier leakage.

\begin{table*}[!h]
\centering
\caption{Statistics of packet numbers and tokenized flow lengths of the dataset.}
\label{flowlenstats}
\resizebox{\textwidth}{!}{
\begin{tabular}{lcccccc}
\hline
\textbf{Dataset} & \textbf{0 percentile} & \textbf{25 percentile} & \textbf{50 percentile} & \textbf{75 percentile} & \textbf{100 percentile} \\
\hline
Number of Packets in each flow & 5 & 10 & 21 & \textbf{34} & 2,429,639 \\
The Length of tokenized flow & 175 & 567 & 973 & \textbf{1,466} & 126,871,264 \\
\hline
\end{tabular}}
\end{table*}

We truncate each flow to its first 34 packets, corresponding to the 75th 
percentile of the packet-length distribution as shown in Table~\ref{flowlenstats}. 
This choice balances (i) retaining sufficient semantic information 
contained in early packets (e.g., handshake, protocol negotiation, 
application identification), and (ii) maintaining a manageable 
context length for Transformer-based models. 
Empirically, using 34 packets preserves $>=75\%$ of classification-relevant 
signals while keeping the sequence length below 1.5k tokens, 
which is the maximum context length supported by TurboT5 on our hardware.

\subsection{Dataset Percentage}\label{appendix:ptftpercent}
\begin{table}[!thb]
    \centering
    \caption{Dataset for Pretraining and Finetuning. The Cross Platform dataset are excluded from pretraining; used for generalization evaluation in finetuning.}
    \label{tab:ptftdist}
    \begin{tabular}{llc}
        \hline
        \textbf{Dataset}& \textbf{Pretraining} & \textbf{Finetuning} \\
        \hline
        ISCX-VPN\cite{vpn} & 60\% &  24\%\\
        ISCX-Tor\cite{tor} & 60\% &  24\%\\
        USTC-TFC-2016\cite{ustc} & 60\% & 5\% \\
        Cross Platform (Android)\cite{cp} & NA & 15\%\\
        Cross Platform (IOS)\cite{cp} & NA & 26\%\\
        CIC-DoHBrw-2020\cite{doh} & 60\% & 6\% \\
        CIC-IoT-2023\cite{iot} & 60\% & 5\% \\
        \hline
    \end{tabular}
\end{table}

\subsection{Benchmark Tasks}\label{appendix:tasks}
\begin{table*}[h]
\centering
\caption{Overview of 12 Classification Tasks Across Datasets}
\label{tab:tasks}
\begin{tabular}{|c|l|l|c|}
\hline
\textbf{Task \#} & \textbf{Dataset} & \textbf{Task Description} & \textbf{Classes} \\
\hline
Task 1 & ISCX-VPN & VPN detection & 2 \\
\hline
Task 2 & ISCX-VPN & VPN Service detection & 6 \\
\hline
Task 3 & ISCX-VPN & VPN application classification & 16 \\
\hline
Task 4 & ISCX-Tor and USTC-TFC-2016 & Tor service detection & 7 \\
\hline
Task 5 & USTC-TFC-2016 & Application Classification & 16 \\
\hline
Task 6 & Cross Platform (Android) & Application classification & 209 \\
\hline
Task 7 & Cross Platform (Android) & Country detection & 3 \\
\hline
Task 8 & Cross Platform (iOS) & Application classification & 196 \\
\hline
Task 9 & Cross Platform (iOS) & Country detection & 3 \\
\hline
Task 10 & CIC-DoHBrw-2020 (DoH) & DoH query method classification & 5 \\
\hline
Task 11 & CIC-IoT-2023 & IoT attack detection & 2 \\
\hline
Task 12 & CIC-IoT-2023 & IoT attack method detection & 7 \\
\hline
\end{tabular}
\end{table*}

\newpage
\section{Ablation and Analysis of Pretraining Objective}
\subsection{Comparison with Random Masking Strategies}\label{appendix:msp}
\textbf{KG-MSP consistently outperforms random masking during pretraining.} As shown in Table~\ref{table:mspablation}, \Lens achieves +1.91\% / +8.6\% higher accuracy and +3.26\% / +4.05\% higher F1 on Task 2 and Task 3, respectively. This improvement stems from masking protocol-critical metadata and payload-related information rather than arbitrary spans. Since random masking here corresponds to the MASS-style/T5-style span masking objective~\cite{t5}, these results highlight the benefit of our knowledge-guided design.

\begin{table}[!htb] 
    \centering
    \caption{Ablation study of pretraining strategy with KG-MSP or Random Masking. Models pretrained with KG-MSP perform better than random masking.}
    \label{table:mspablation}
    \begin{tabular}{lcccc}
        \toprule
        \multirow{2}{*}{\centering Pretraining} & \multicolumn{2}{c}{Task 2} & 
        \multicolumn{2}{c}{Task 3} \\
        \cmidrule(lr){2-5}
         & AC & F1 & AC & F1\\
        \midrule
        Random Masking & 0.8788  & 0.8567  & 0.7546 & 0.7732 \\
        KG-MSP(Ours) & \textbf{0.8979} & \textbf{0.8893} & \textbf{0.8406} & \textbf{0.8137} \\
        \bottomrule
    \end{tabular}
\end{table}
\subsection{Sensitivity Analysis of Masking Configuration}\label{appendix:sensitive}
Table~\ref{tab:sensitive_analysis} shows that \Lens achieves its best validation performance when the masking ratio $\theta$ is 60\%, and overall remains stable across different ratios.
Similarly, Table~\ref{tab:sense_seq} indicates that masking Seq/Ack/Length with ratio $k$ also leads to minimal performance variation. This insensitivity arises because KG-MSP defaults to random masking when fewer than 15\% of tokens are masked. For generality, we therefore adopt a 50\% masking ratio.

\begin{table*}[!thb]
    \centering
    \small
    \caption{Sensitivity Analysis on Masking Vital Network Metadata (left) and Seq/Ack/Length (right).}
    \label{tab:sensitive_both}
    \begin{subtable}[t]{0.48\textwidth}
        \centering
        \caption{Analysis of $\theta$ for Masking Vital Metadata}
        \label{tab:sensitive_analysis}
        \begin{tabular}{lccc}
            \toprule
            \multirow{2}{*}{Mask Prob.} &
            30\% & 60\% & 90\% \\
            \cmidrule(lr){2-4}
            & AC & AC & AC \\
            \midrule
            \Lens~(Ours) & 0.33 & \textbf{0.34} & 0.32 \\
            \bottomrule
        \end{tabular}
    \end{subtable}
    \hfill
    \begin{subtable}[t]{0.48\textwidth}
        \centering
        \caption{Analysis of $k$ for Masking Seq/Ack/Length}
        \label{tab:sense_seq}
        \begin{tabular}{lccc}
            \toprule
            \multirow{2}{*}{Mask Prob.} &
            25\% & 50\% & 75\% \\
            \cmidrule(lr){2-4}
            & AC & AC & AC \\
            \midrule
            \Lens~(Ours) & \textbf{0.35} & \textbf{0.35} & 0.34 \\
            \bottomrule
        \end{tabular}
    \end{subtable}
\end{table*}

\newpage
\section{More detail on Experimental Results}
\subsection{Performance of per-class in the extensibility of classification}\label{appendix:extensibility}
We summarize the setup of the extensibility experiments. For Task 6, the new classes are selected based on their higher sample counts: com.ifeng.news2 (226) for the 1-class setting; additionally, sohu.sohuvideo (200) and xunlei.downloadprovider (190) for the 3-class setting; and further qiyi.video (150) and youku.phone (150) for the 5-class setting.
For Task 8, we follow the same criterion: pocket-pool (416) for the 1-class setting; plus aiqiyi (238) and color-ballz (224) for the 3-class setting; and further yy (206) and youku (182) for the 5-class setting. These labels provide clearer performance differences due to their larger test sets.

Across Table~\ref{tab:1novelclass}, \ref{tab:3novelclass}, and \ref{tab:5novelclass}, \Lens consistently outperforms all baselines on both tasks. ET-BERT fine-tunes new classes with a learning rate of 2e-5, whereas \Lens uses 5e-6 due to architectural differences, illustrating \Lens’s stronger extensibility and reduced need for task-specific adjustment.

\begin{table}[htb]
    \centering
    \caption{Performance of 1 new class in Task 6 and Task 8.}
    \label{tab:1novelclass}
    \begin{subtable}[t]{0.47\linewidth}
    \centering
    \caption{Task 6 — com.ifeng.news2}
    \begin{tabular}{lccc}
        \toprule
        Methods & P & R & F1 \\
        \midrule
        ET-BERT & 0.0474 & \textbf{1.0000} & 0.0900 \\
        \textbf{Lens} & \textbf{0.9636} & 0.9408 & \textbf{0.9521} \\
        \bottomrule
    \end{tabular}
    \end{subtable}
    \hfill
    \begin{subtable}[t]{0.47\linewidth}
    \centering
    \caption{Task 8 — pocket-pool}
    \begin{tabular}{lccc}
        \toprule
        Methods & P & R & F1 \\
        \midrule
        ET-BERT & 0.0643 & \textbf{1.0000} & 0.1208 \\
        \textbf{Lens} & \textbf{0.6753} & \textbf{1.0000} & \textbf{0.8062} \\
        \bottomrule
    \end{tabular}
    \end{subtable}

\end{table}

\begin{table*}[htb]
\centering
\caption{Performance of 3 new classes in Task 6 and Task 8.}
\label{tab:3novelclass}
\vspace{-0.05in}

\subcaption*{Task 6}
\begin{tabular}{lccccccccc}
\toprule
Methods & \multicolumn{3}{c}{com.ifeng.news2} 
        & \multicolumn{3}{c}{sohu.sohuvideo} 
        & \multicolumn{3}{c}{xunlei.downloadprovider} \\
\cmidrule(lr){2-10}
 & P & R & F1 & P & R & F1 & P & R & F1 \\
\midrule
ET-BERT     & 0.0484 & \textbf{0.9941} & 0.0924
            & 0.2213 & \textbf{0.9858} & 0.3615
            & 0.5930 & 0.9440 & 0.7284 \\
\textbf{Lens} 
            & \textbf{0.8195} & \textbf{0.9941} & \textbf{0.8984}
            & \textbf{0.9577} & 0.9645 & \textbf{0.9611}
            & \textbf{0.9191} & \textbf{1.0000} & \textbf{0.9579} \\
\bottomrule
\end{tabular}
\vspace{0.15in}

\subcaption*{Task 8}
\begin{tabular}{lccccccccc}
\toprule
Methods & \multicolumn{3}{c}{pocket-pool} 
        & \multicolumn{3}{c}{aiqiyi}
        & \multicolumn{3}{c}{color-ballz} \\
\cmidrule(lr){2-10}
 & P & R & F1 & P & R & F1 & P & R & F1 \\
\midrule
ET-BERT 
        & 0.1098 & \textbf{1.0000} & 0.1978
        & 0.0995 & 0.9888 & 0.1809
        & 0.5685 & 0.9881 & 0.7217 \\
\textbf{Lens}
        & \textbf{0.8041} & \textbf{1.0000} & \textbf{0.8914}
        & \textbf{0.4120} & \textbf{1.0000} & \textbf{0.5836}
        & \textbf{0.7000} & \textbf{1.0000} & \textbf{0.8235} \\
\bottomrule
\end{tabular}
\end{table*}
\begin{table*}[htb]
\centering
\caption{Performance of 5 new classes in Task 6 and Task 8.}
\label{tab:5novelclass}
\vspace{-0.05in}

\subcaption*{Task 6}
\resizebox{\textwidth}{!}{
\begin{tabular}{lccccccccccccccc}
\toprule
Methods 
& \multicolumn{3}{c}{com.ifeng.news2}
& \multicolumn{3}{c}{sohu.sohuvideo}
& \multicolumn{3}{c}{xunlei.downloadprovider}
& \multicolumn{3}{c}{qiyi.video}
& \multicolumn{3}{c}{youku.phone} \\
\cmidrule(lr){2-16}
 & P & R & F1 & P & R & F1 & P & R & F1 & P & R & F1 & P & R & F1 \\
\midrule
ET-BERT 
& 0.0572 & \textbf{1.0000} & 0.1083
& 0.2945 & \textbf{0.9858} & 0.4535
& 0.4237 & \textbf{1.0000} & 0.5952
& 0.3839 & \textbf{1.0000} & 0.5548
& 0.5170 & 0.9681 & 0.6741 \\
\textbf{Lens}
& \textbf{0.7249} & 0.9822 & \textbf{0.8342}
& \textbf{0.8528} & \textbf{0.9858} & \textbf{0.9145}
& \textbf{0.8389} & \textbf{1.0000} & \textbf{0.9124}
& \textbf{0.7532} & \textbf{1.0000} & \textbf{0.8592}
& \textbf{0.6462} & 0.8936 & \textbf{0.7500} \\
\bottomrule
\end{tabular}
}
\vspace{0.15in}

\subcaption*{Task 8}
\resizebox{\textwidth}{!}{
\begin{tabular}{lccccccccccccccc}
\toprule
Methods 
& \multicolumn{3}{c}{pocket-pool}
& \multicolumn{3}{c}{aiqiyi}
& \multicolumn{3}{c}{color-ballz}
& \multicolumn{3}{c}{yy}
& \multicolumn{3}{c}{youku} \\
\cmidrule(lr){2-16}
 & P & R & F1 & P & R & F1 & P & R & F1 & P & R & F1 & P & R & F1 \\
\midrule
ET-BERT 
& 0.2188 & \textbf{1.0000} & 0.3590
& 0.2607 & 0.9551 & 0.4096
& \textbf{0.6484} & 0.9881 & \textbf{0.7830}
& 0.4444 & 0.9870 & 0.6129
& 0.0000 & 0.0000 & 0.0000 \\
\textbf{Lens}
& \textbf{0.8254} & \textbf{1.0000} & \textbf{0.9043}
& \textbf{0.5733} & \textbf{0.9663} & \textbf{0.7197}
& 0.6336 & \textbf{0.9881} & 0.7721
& \textbf{0.5385} & \textbf{1.0000} & \textbf{0.7000}
& \textbf{0.6476} & \textbf{1.0000} & \textbf{0.7861} \\
\bottomrule
\end{tabular}
}
\end{table*}

\newpage
\section{Examples of Model Input}\label{appendix:input_example}
\begin{figure*}[!thb]
    \centering
    \includegraphics[width=\textwidth]{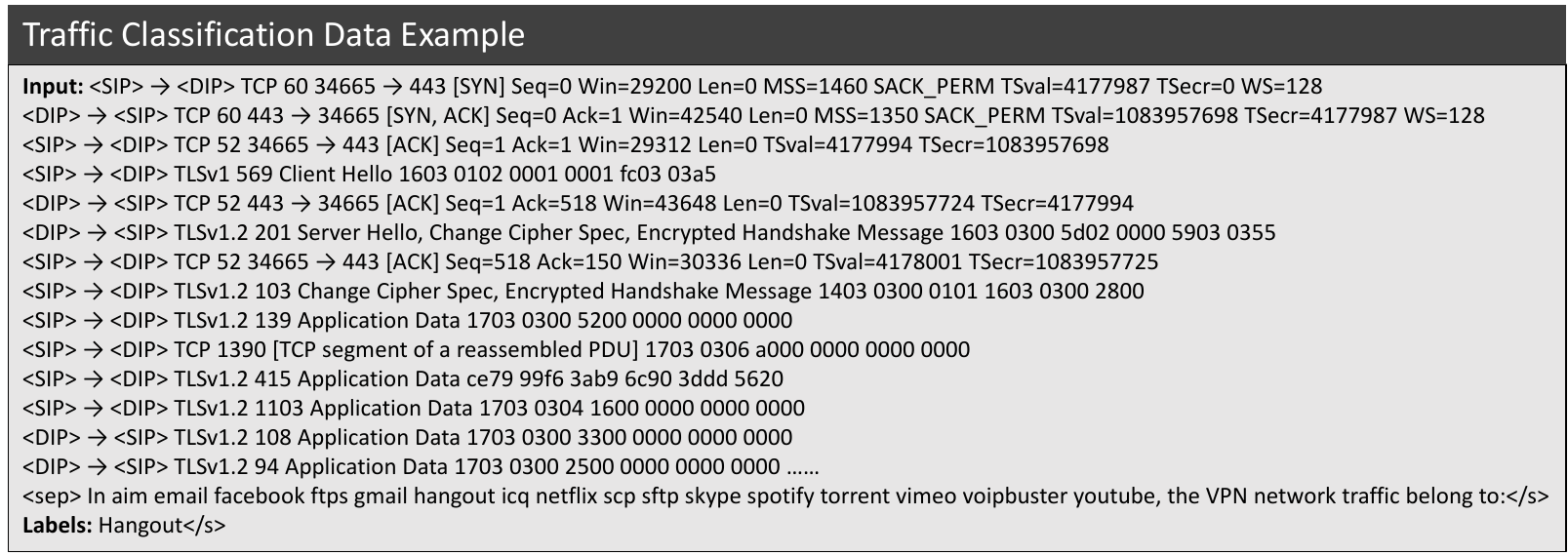}
    \includegraphics[width=\textwidth]{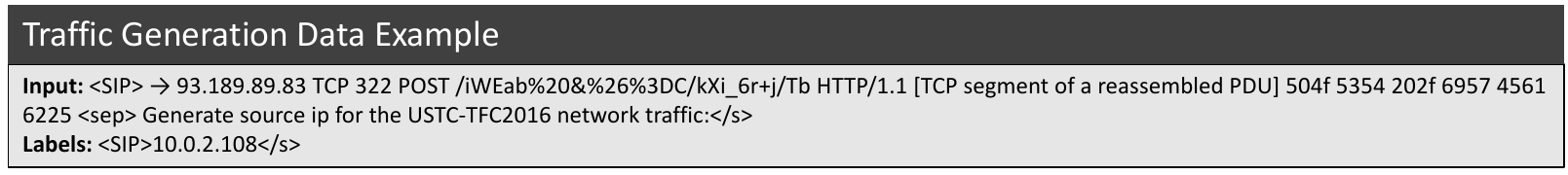}
    \caption{The example input of classification and generation. For classification, the input includes parsed network traffic and a task context listing label options. For generation, the input contains a masked packet and a task description.}
    \label{fig:dataexample}
\end{figure*}

\section{Hyperparameter Setting}\label{appendix:hyper}
\begin{table*}[!thb]
\centering
\caption{Hyperparameters for \Lens.}
\label{tab:hyperparameters}
\begin{tabular}{llc}
\hline
& \textbf{Hyperparameter} & \textbf{Value} \\
\hline
& Transformer Encoder Layer number & 12 \\
& Transformer Decoder Layer number & 12 \\
& Attention heads number & 12 \\
Architecture & Attention heads dimension & 64 \\
& Hidden dimension & 768 \\
& MLP hidden & 2048 \\
& MLP activation & Gated-GeLU \\
\hline
& Total gradient steps & 780k \\
& Batch size & 48 \\
& Learning rate & $1 \times 5^{-4}$ \\
Pre-training & Dropout rate & 0.1 \\
& Optimizer & AdamW \\
& Scheduler & Warmup cosine \\
& Grad clip norm & $1.0$ \\
& Scheduler warmup steps & 13k \\
\hline
& Total Max epochs & 40 \\
& Batch size & 32 \\
Fine-tuning & Learning rate & $1 \times 5^{-5}$ \\
& Dropout rate & 0.1 \\
& Optimizer & AdamW \\
& Grad clip norm & $1.0$ \\
& Scheduler & Warmup linear \\
\hline
\end{tabular}
\end{table*}

\end{document}